\documentclass[article]{elsarticle}

\usepackage{natbib}
\usepackage{geometry}
\usepackage{graphicx}
\usepackage{subfigure}
\usepackage{scrtime}
\usepackage{epstopdf}
\usepackage{amsmath}
\usepackage{siunitx}
\usepackage{amssymb}
\usepackage{upgreek}
\usepackage{appendix}
\usepackage[colorlinks, linkcolor=black, citecolor=black, anchorcolor=black]{hyperref}
\hypersetup{hidelinks}
\usepackage{algorithm}
\usepackage{algorithmicx}
\usepackage{algpseudocode}

\usepackage{multirow}
\usepackage{lineno}
\usepackage{enumerate}
\usepackage[section]{placeins}
\usepackage{csquotes}

\usepackage{threeparttable}

\usepackage{booktabs}
\usepackage[dvipsnames]{xcolor}
\usepackage{tcolorbox}
\usepackage{soul}
\usepackage{makecell}
\usepackage{amsthm}
\usepackage{lscape}
\usepackage{mathtools}
\theoremstyle{remark}
\newtheorem{remark}{Remark}

\def\romannumer#1{\uppercase\expandafter{\romannumeral#1}}

\makeatletter
\def\ps@pprintTitle{
	\let\@oddhead\@empty
	\let\@evenhead\@empty
	\def\@oddfoot{\centerline{\thepage}}%
	\let\@evenfoot\@oddfoot}
\makeatother

\begin{document}

\begin{frontmatter}
\title{ArGEnT: Arbitrary Geometry-encoded Transformer for Operator Learning}
\author[1]{Wenqian Chen }
\author[1]{Zhi-Feng Wei }
\author[1]{Yucheng Fu \corref{cor1}}
\ead{yucheng.fu@pnnl.gov}
\author[2]{Michael Penwarden}
\author[3]{Pratanu Roy}
\author[1]{Panos Stinis}


\cortext[cor1]{Corresponding author}

\address[1]{Pacific Northwest National Laboratory, Richland, WA 99352, USA}
\address[2]{Sandia National Laboratories, Albuquerque, NM 87123, USA}
\address[3]{Lawrence Livermore National Laboratory, Livermore, CA 94550, USA}

\begin{abstract}
Learning solution operators on arbitrary geometries remains a central challenge in scientific machine learning. This challenge is increasingly important for many-query simulation, physics-informed learning, and adaptation to evolving geometries, where predictions must be accurate, geometry-aware, and evaluable at arbitrary spatial locations. Existing operator-learning methods often rely on structured discretizations, explicit geometry parameterizations, or point-cloud formulations that couple geometric representation with solution-query sampling, limiting their flexibility for irregular and non-parameterized domains. In this work, we propose the Arbitrary Geometry-encoded Transformer (ArGEnT), a geometry-conditioned attention framework that decouples geometry encoding from query-point evaluation. We develop three ArGEnT variants: self-attention, cross-attention, and hybrid-attention. Importantly, ArGEnT is a standalone geometry-query encoding architecture that can be used independently or integrated with different neural operators to incorporate non-geometric physical inputs. In the cross-attention variant, geometry is represented by an independently sampled point cloud used to construct keys and values, while arbitrary solution-query points construct queries. This design enables mesh-independent field prediction, reduces sensitivity to the query-point distribution, and allows compact geometric representations to condition large-scale solution evaluations. Across benchmark problems in fluid dynamics, solid mechanics, and electrochemical systems, ArGEnT consistently improves accuracy and generalization over standard DeepONet, point-cloud-based operator learning, and geometry-aware transformer baselines. In several cases, ArGEnT reduces prediction errors by more than an order of magnitude while requiring substantially lower training cost than transformer-based baselines. These results demonstrate that decoupled geometry-query attention provides an accurate, scalable, and flexible neural operator-learning framework for arbitrary geometries.
\end{abstract}


\end{frontmatter}

\section{Introduction}
Surrogate modeling has become a fundamental tool for accelerating scientific computation and engineering analysis by providing efficient approximations to computationally expensive numerical simulations~\cite{benner2015survey}. By replacing repeated high-fidelity solver calls with fast learned evaluations, surrogate models can reduce computational cost by orders of magnitude while enabling real-time prediction, rapid parametric studies, and large-scale optimization. In many practical settings, both the geometry of the computational domain and the governing operators vary across problem instances, as in geometry optimization for fluid flow~\cite{jameson2003aerodynamic, sun2023physics}, structural response under changing configurations~\cite{sokolowski1992introduction, samadian2025application}, and multiphysics systems with varying boundary conditions or source terms~\cite{wang2023integration, li2024maximizing}. Surrogate models that can accommodate both arbitrary geometries and varying operators eliminate the need for retraining or remeshing for every new configuration and improve data efficiency by reusing learned representations across diverse scenarios. These advantages are especially attractive in high-dimensional, many-query applications, where conventional numerical solvers become prohibitively expensive.

Early neural-network-based surrogate models \cite{hesthaven2018non, wang2019non} relied primarily on multilayer perceptrons (MLPs) to approximate mappings between fixed-dimensional inputs and outputs. Although MLPs are universal function approximators \cite{HORNIK1989359}, their practical performance depends strongly on carefully designed parameterizations of geometry and physical inputs. Convolutional neural networks (CNNs) extended this paradigm by exploiting spatial locality and translation invariance, making them effective for surrogate modeling on structured grids and image-like field representations. CNN-based surrogates have been widely used in applications such as fluid dynamics\cite{zhang2025convolutional}, heat transfer \cite{hua2023surrogate}, and porous media flow \cite{kim2025prt}, but their reliance on regular grids limits their ability to handle complex or irregular geometries directly.

To move beyond fixed input-output mappings, operator-learning methods have been developed to learn mappings between function spaces. Deep Operator Networks (DeepONets) \cite{lu2021learning} use branch-trunk architectures to represent operators explicitly and have shown strong performance in learning partial differential equation (PDE) solution operators \cite{he2024geom, peyvan2025fusion, SHUKLA2024107615}. Fourier Neural Operators (FNOs) \cite{li2020fourier} {perform convolutions in the spectral domain on uniform grids, enabling efficient modeling of global spatial interactions. These methods substantially improve operator-level generalization, but they typically still require structured discretizations or explicit geometry encodings}\cite{li2023fourier, bonev2023spherical}.

Geometry-aware neural architectures have been developed to address arbitrary and irregular domains more effectively. Graph neural networks (GNNs) naturally operate on unstructured meshes and graph representations, making them well suited for surrogate modeling on complex geometries \cite{wu2020comprehensive,horie2022physics}. Point-based networks, such as PointNets \cite{qi2017pointnet, kashefi2022physics} and their variant Point-DeepONet\cite{park2026point}, process point clouds directly without requiring mesh connectivity and therefore provide a flexible representation for irregular domains. Transformer\cite{vaswani2017attention}-based models \cite{li2022transformer,wen2025geometry,liu2026geometry, alkin2025ab}, equipped with attention mechanisms, offer a powerful framework for capturing long-range dependencies and integrating geometric information with physical parameters, and have shown strong promise for learning operators across varying geometries.

{In this work, we propose ArGEnT, an Arbitrary Geometry-encoded Transformer, as a geometry-aware encoder for operator learning on complex and irregular domains. ArGEnT develops a unified family of three attention formulations—self-attention, cross-attention, and hybrid-attention—to systematically investigate different mechanisms for encoding geometry and transferring geometric information to solution representations. ArGEnT is partially inspired by OFormer~\cite{li2022transformer}, whose encoder-query formulation first uses self-attention to encode the input/domain representation and then uses cross-attention to connect that encoded representation with arbitrary query points. OFormer therefore should not be characterized as relying solely on self-attention. Related geometry-aware operator approaches such as GAOT~\cite{wen2025geometry} use self-attention for geometry or domain encoding. AB-UPT \cite{alkin2025ab}  adopts a CFD-oriented multi-branch architecture: geometry is compressed into latent geometry tokens, which condition separate surface and volume physics branches through cross-attention, followed by additional cross-attention interactions between the surface and volume physical representations. ArGEnT instead directly formulates the interaction between geometry points and arbitrary solution-query locations, without requiring separate surface and volume physics branches; this is an architectural distinction and a difference in modeling objective, rather than a claim of universal superiority.} {The cross-attention formulation introduces a decoupled geometry-query representation: independently sampled geometry points construct the keys and values, while arbitrary solution-query points construct the queries. In the pure cross-attention formulation, query points do not interact with one another. Once the geometry representation is fixed, each query point is propagated and evaluated independently, enabling flexible query sampling, arbitrary mini-batching or chunked evaluation, and predictions that do not depend on which other query points are included in the same forward pass.} ArGEnT is a standalone, architecture-agnostic geometry-query encoding framework that can operate independently or be integrated with different operator-learning frameworks when additional non-geometric physical inputs are required. {Here, we integrate the geometry-query encoder into an operator-learning framework through a DeepONet-style formulation. ArGEnT provides the geometry-aware trunk representation for geometry and solution-query locations, while the branch network independently encodes non-geometric physical or operating parameters. This separation assigns geometry-dependent and non-geometric dependence to dedicated components and does not require geometry to be explicitly reduced to a low-dimensional parameterization in the branch input. The use of DeepONet itself is not the methodological novelty; rather, it provides a representative operator-learning interface for the proposed geometry-query encoder.}

\section{Methodology}
\subsection{Arbitrary Geometry-encoded Transformer (ArGEnT)}
\begin{figure}[htbp]
    \centering
    \includegraphics[width=12cm, trim=0cm 0cm 0cm 0cm, clip=true]{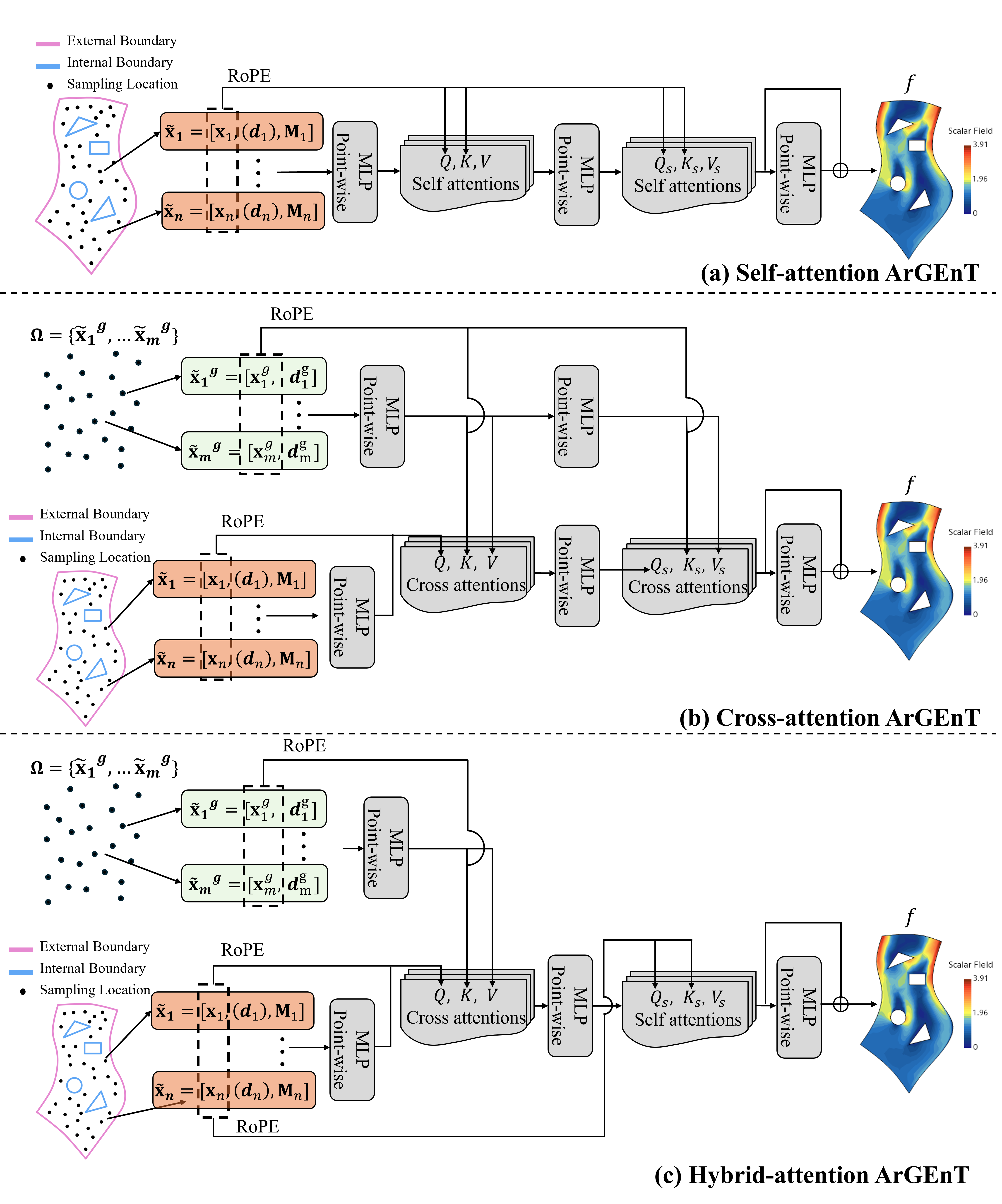}
    \caption{Arbitrary Geometry-encoded Transformer (ArGEnT). (a) Two-layer self-attention transformer; (b) two-layer cross-attention transformer; (c) hybrid-attention transformer composed of one cross-attention layer followed by one self-attention layer. $\mathbf{x}$ denotes the input point coordinates; $M$ is the boolean mask indicating padding points; $d$ represents the signed distance function (SDF) values, where $(\cdot)$ indicates that including SDF inputs are optional. Q, K, and V denote the query, key, and value matrices in the attention mechanism. RoPE denotes Rotary Position Embeddings used to incorporate relative positional information from the input coordinates. MLP refers to multi-layer perceptron layers. $\oplus$ indicates a residual connection.}
    \label{Fig_transformer_architectures}
\end{figure}

For problems involving complex geometries, traditional MLP-based neural networks often struggle to represent geometric structure effectively and to capture the corresponding geometry-dependent solutions. To address this limitation, we introduce a family of geometry-encoded transformer architectures that exploit the attention mechanism to capture intricate geometric features and their influence on the target function. Attention enables the model to learn long-range dependencies and complex relationships within the input data, which is especially valuable when the geometry is difficult to parameterize explicitly. Based on how geometric information enters the attention operation, we classify the proposed architectures into three variants: self-attention, cross-attention, and hybrid-attention, as illustrated in Figure \ref{Fig_transformer_architectures}.


In the self-attention transformer shown in Fig. \ref{Fig_transformer_architectures}(a), geometric information is implicitly encoded in the input data itself. The self-attention mechanism uses the spatial coordinates of a point cloud together with associated features, if any, to construct the query (Q), key (K), and value (V) matrices, as detailed in \ref{appendix_self_attention}. Because the point distribution typically comes from simulation data, it already contains geometric information about the shape and structure of the domain. The input is organized as a three-dimensional tensor with batch, point, and feature dimensions. The feature dimension includes the spatial coordinates, optional SDF values, and mask values. The coordinates provide positional information for each point, while the SDF can be included to provide additional geometric context; the use of SDF values as additional input features is discussed in \mbox{\ref{sec_SDF}}. {Mask values distinguish valid points from padded points. Because the number of points varies across geometries, shorter point sets are extended with dummy points to obtain a consistent input size within each batch. A binary mask is created for each point set, where 1 indicates a valid point and 0 indicates a padded point. In the linear-attention layer as shown in \ref{appendix_self_attention} and \ref{appendix_cross_attention}, the mask is multiplied elementwise with the key and value tensors, setting the representations of padded points to zero. Consequently, padded points do not contribute to the attention aggregation. The aggregated features are then normalized by the number of valid points, obtained by summing the mask, rather than by the total padded sequence length.}

Before the tensor is passed to the attention layers, an MLP projects the input features to a higher-dimensional space so that the attention mechanism can capture more complex relationships. Rotary Position Embeddings (RoPE), described in \ref{appendix_RoPE}, are then applied to the query and key matrices to encode relative positional information based on the spatial coordinates. This allows the attention block to account explicitly for geometric distance when computing attention scores. The output of each self-attention layer is fed recursively into the next layer, and the output of the final layer is processed by additional MLP layers within a residual block to produce the final output.

In the cross-attention transformer shown in Fig. \ref{Fig_transformer_architectures}(b), geometric information is supplied as a separate input consisting of the spatial coordinates of a fixed point cloud and their corresponding SDF values. The point cloud is sampled once within a finite spatial domain and then kept identical across all cases to eliminate sampling-induced variation. As a result, geometric variability is encoded entirely through the SDF field defined on this fixed set of spatial locations. The geometric input is used to construct the key and value matrices in the attention mechanism, while the main input data, such as spatial coordinates and other features, is used to construct the query matrix, as detailed in \ref{appendix_cross_attention}. This decoupling allows the model to process query data independently of the geometry representation, which is especially useful when predictions are needed at arbitrary spatial locations because the query points can be sampled independently of the geometry. As in the self-attention transformer, MLP layers first project both the query input and the geometric input to higher-dimensional feature spaces. RoPE is then applied to the query and key matrices to encode relative positional information based on the spatial coordinates. The output of each cross-attention layer is passed recursively to the next layer as the updated query input, and the output of the final layer is processed by additional MLP layers within a residual block to produce the final output.

Because self-attention captures geometry implicitly from the input-point distribution, it can potentially learn relationships that are not explicitly encoded in the geometric input. Cross-attention, by contrast, provides explicit geometric encoding, which can help the model focus on relevant geometric features when processing the query data. To combine the strengths of both approaches, we also propose a hybrid-attention transformer architecture that uses both cross-attention and self-attention. As shown in Fig. \ref{Fig_transformer_architectures}(c), the hybrid-attention transformer consists of an initial cross-attention layer followed by a self-attention layer. As in the cross-attention transformer, the geometric input is used to construct the key and value matrices in the first layer, while the main input data is used to construct the query matrix. The output of the cross-attention layer is then passed to the self-attention layer, which further captures relationships and dependencies within the geometric context. This combination allows the model to benefit from both implicit and explicit geometric encoding and can therefore improve performance on complex geometry-dependent features. Because the hybrid-attention transformer begins with a cross-attention layer, it also allows the query points to be sampled independently of the geometry representation, similar to the cross-attention transformer. Unless otherwise specified, the hybrid-attention transformer uses the same input setup as the cross-attention transformer.

Overall, the learned operator for the ArGEnT models can be expressed as:
\begin{equation}\label{eq_ArGEnT_trunk}
\begin{aligned}
\textbf{Self-attention} \quad  \mathcal{G} &:  \left\{\tilde{\mathbf{x}}_i\right \}_{i=1}^{N}
&\longmapsto& \{f_i\}_{i=1}^{N}  \\  
\textbf{Cross-attention} \quad \mathcal{G} &: (\tilde{\mathbf{x}}, \Omega) 
&\longmapsto& f  \\
\textbf{Hybrid-attention} \quad \mathcal{G} &:   (\left\{\tilde{\mathbf{x}}_i\right \}_{i=1}^{N}, \Omega) 
&\longmapsto& \{f_i\}_{i=1}^{N}
\end{aligned}
\end{equation}
where $\tilde{\mathbf{x}} =(\mathbf{x}, d)$ denotes the query points, $\mathbf{x} \in \mathbb{R}^2$ (or $\mathbb{R}^3$) the spatial coordinates, $d$ the optional SDF value, and other features can be appended when needed. $\Omega$ denotes the geometry and it is represented as a point cloud $\Omega:=\{\tilde{\mathbf{x}}_i^g \}_{i=1}^{M}$, where $\tilde{\mathbf{x}}^g =(\mathbf{x}^g, d^g)$ denotes a geometry point with its SDF value. Note that for the self- and hybrid-attention ArGEnT models, the learned operator maps a set of input points to a set of output values, rather than performing a point-wise (point-to-point) mapping as in the cross-attention ArGEnT. This distinction arises because self-attention explicitly exploits the mutual relationships among input points to construct the attention representation. Consequently, the sampling strategy for the input points can influence the model predictions, an effect that is discussed in more detail in Section~\ref{sec_airfoil}. In contrast, the cross-attention ArGEnT decouples the query points from the geometry representation, allowing them to be sampled independently and enabling flexible evaluation at arbitrary spatial locations. {More precisely, for a fixed geometry, the geometry-derived keys and values remain fixed, and each solution-query point is propagated independently through the cross-attention operation. Pure cross-attention contains no interaction among query points; therefore, the prediction at one query point does not depend on the number, distribution, or ordering of other query points included in the same forward pass. The same geometry representation can consequently be reused for arbitrary query mini-batches or chunked evaluations.}

\subsection{ArGEnT DeepONet Architecture}
\begin{figure}[htbp]
    \centering
    \includegraphics[width=10cm, trim=0cm 0cm 0cm 0cm, clip=true]{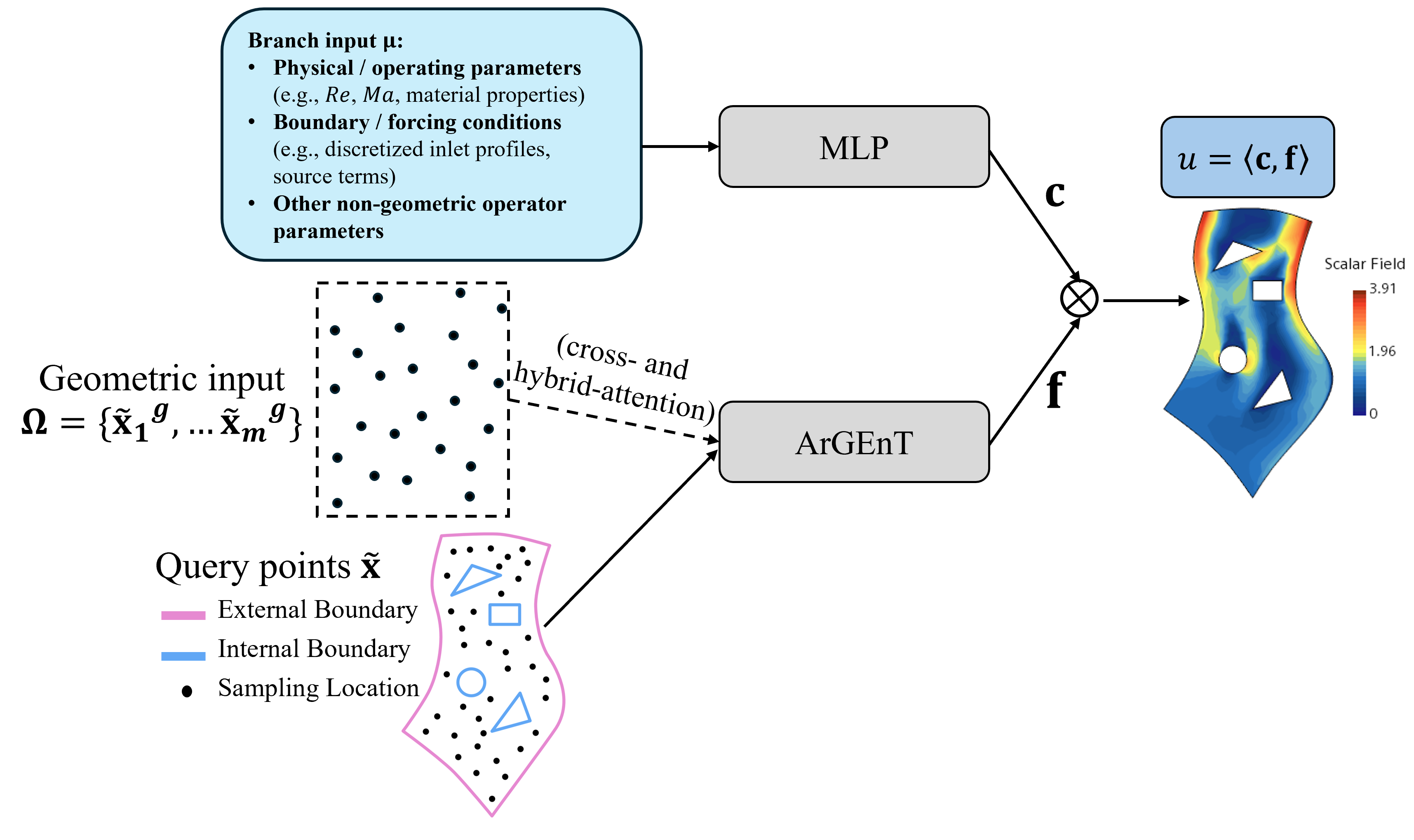}
    \caption{ArGEnT DeepONet architecture. The ArGEnT model functions as the trunk network, responsible for encoding geometric representations and query information, whereas the branch network processes non-geometric input parameters. The final prediction of the target function is obtained by taking the inner product of the trunk and branch outputs.}
    \label{Fig_ArGEnT_DeepONet}
\end{figure}

{ArGEnT first provides a geometry-query representation: it encodes the geometry and solution-query locations and thereby acts as the geometry-aware trunk representation. To incorporate additional non-geometric inputs, we integrate this encoder into an operator-learning framework through the Deep Operator Network (DeepONet) formulation~\cite{lu2021learning}, yielding the ArGEnT DeepONet architecture shown in Fig.~\ref{Fig_ArGEnT_DeepONet}. The branch network independently encodes physical or operating parameters, such as boundary conditions and material properties, while the ArGEnT trunk represents the geometry-dependent query features. This separation allows geometry-dependent and non-geometric dependence to be represented through dedicated components, without requiring geometry to be explicitly reduced to a low-dimensional parameterization in the branch input. When multiple non-geometric input parameters are present, each set of parameters can be processed through a separate branch network, as in MIONet~\cite{jin2022mionet}. The methodological contribution is the integration of the proposed geometry-query encoder into operator learning, rather than the use of DeepONet itself.}

The learned operator for the ArGEnT DeepONet can be expressed as:
{\small
\begin{equation}\label{eq_ArGEnT}
\begin{aligned}
\textbf{Self-attention} \quad
\mathcal{G} &:\; (\{\tilde{\mathbf{x}}_i\}_{i=1}^N, \pmb{\mu})
\;&\longmapsto&\;
\left\{
u_i = \sum_{j=1}^J c^j(\pmb{\mu})\,
f_i^j\!\left(\{\tilde{\mathbf{x}}_k\}_{k=1}^N\right)
\right\}_{i=1}^N  \\
\textbf{Cross-attention} \quad
\quad \mathcal{G} &: \quad (\tilde{\mathbf{x}}, \Omega, \pmb{\mu}) 
\;&\longmapsto&\;
u = \sum_{j=1}^J c^j(\pmb{\mu})\,
f^j\!\left(\tilde{\mathbf{x}}, \Omega\right)
 \\
\textbf{Hybrid-attention} \quad
\mathcal{G} &:\; (\{\tilde{\mathbf{x}}_i\}_{i=1}^N, \Omega, \pmb{\mu})
\;&\longmapsto&\;
\left\{
u_i = \sum_{j=1}^J c^j(\pmb{\mu})\,
f_i^j\!\left(\{\tilde{\mathbf{x}}_k\}_{k=1}^N, \Omega\right)
\right\}_{i=1}^N 
\end{aligned}
\end{equation}
}
where J is the output dimension of trunk and branch networks, $c^j$ is the $j$th output of branch network, and $f_i^j$ is the $j$th output of trunk network evaluated at the $i$th point. $\pmb{\mu}$ denotes the non-geometric input parameters processed by the branch network. $\pmb{\mu}$ can be physical/operating parameters, boundary/initial conditions represented as functions,  or  other non-geometric parameters.

\subsection{Training setup}
Because the number of training points can be large, sometimes exceeding 100,000, we adopt a mini-batch training strategy. At each training step, a random subset of 3,000 query points is sampled from the full training set to evaluate the loss function and update the model parameters. This strategy significantly reduces memory consumption and improves training efficiency. All models are implemented in PyTorch and trained on NVIDIA H100 GPUs. Unless stated otherwise, all transformer-based models are optimized using Adam for a total of 100,000 training steps. The learning rate is initialized at 0.001 and decays by a factor of 0.99 every 200 training steps. The mean squared error (MSE) loss is used to quantify the discrepancy between predicted and ground-truth target values during training.
For evaluation, we use relative $L^2$ error by default; for benchmark comparisons, however, we follow the error metrics reported in prior work, using MSE for the turbulent airfoil problem in Section \ref{sec_turbulent_airfoil}  and mean absolute error (MAE) for the jet engine bracket problem in Section \ref{sec_bracket}.

\begin{remark}
\textit{For all benchmark comparisons, Point-DeepONet, Transolver, and Transolver+ are implemented using the architectures and official open-source repositories provided in the original works\cite{park2026point, wu2024transolver, luo2025transolver++}. Aside from minor data-specific preprocessing and necessary training-script adjustments, we use the default or recommended hyperparameters, unless otherwise noted, without additional tuning.}
\end{remark}

The mini-batch training strategy can degrade the accuracy of self-attention and hybrid-attention transformers because randomly sampled query points in each mini-batch may not follow the same distribution as the full training set. This mismatch can adversely affect the self-attention mechanism, which relies on global context. Although the issue can be mitigated partially by increasing the mini-batch size, doing so significantly raises memory consumption and computational cost, thereby limiting the scalability of self-attention mechanisms for large-scale problems or complex geometries.
In contrast, the cross-attention transformer is largely unaffected because the query points can be sampled independently of the geometry representation. The point cloud used to represent the geometry can therefore remain relatively small while still capturing the essential geometric features without incurring excessive computational cost or accuracy loss. Except for extremely complex geometries, such as porous media or fractal structures, where a larger number of points may be required, this compact representation is sufficient in most practical settings.
Throughout this work, we use 2,000–5,000 geometry points to represent the geometry in both the cross-attention and hybrid-attention transformer models, and the full batch of geometry points is used at each training step. This design makes the cross-attention transformer more scalable and computationally efficient for large-scale problems with complex geometries.

Throughout this work, we use the following hyperparameters for all ArGEnT models unless specified otherwise: 4 attention heads, 128-dimensional features in the attention layers. For the pointwise MLP to process the input features, we use 4 hidden layers each with 128 neurons, and ReLU as the activation function. For the output MLP, we use 3 hidden layers each with 128 neurons, and ReLU as the activation function.
For other pointwise MLP, we do not use any hidden layers, and only use a single linear layer to project the features to the desired output dimension, activated with a $tanh$ function. No dropout or regularization is applied during training. The number of transformer layers is set to 2 for both self-attention and cross-attention transformers. For the hybrid-attention transformer, we use one cross-attention layer followed by one self-attention layer. We have experimented with deeper transformer architectures (up to 4 layers) but did not observe significant performance improvements, likely due to the relatively small size of the training datasets  in our experiments.

\section{Results and Discussion}
\subsection{Airfoil flow}\label{sec_airfoil}
\subsubsection{Laminar flow over airfoil of varying shapes}\label{sec_laminar_airfoil}
We first consider laminar flow over airfoils with varying profiles at fixed Reynolds number, Mach number, and angle of attack. The dataset, provided by \cite{SHUKLA2024107615}, contains 50 airfoil shapes parameterized by two geometric variables: the maximum camber and the location of maximum camber. The flow conditions are identical for all samples, with Reynolds number 500, Mach number 0.5, and angle of attack $0^\circ$. The flow fields are computed using a high-fidelity CFD solver, and the resulting velocity $(u,v)$, pressure $p$, and density $\rho$ fields serve as the reference solutions for training and evaluation. Further details on dataset generation and numerical setup are given in \cite{SHUKLA2024107615}.

The learned operator for the following airfoil problem, taking the cross-attention ArGEnT as an example,  is therefore defined as
\begin{align}
\textbf{Cross-attention} \quad \mathcal{G} : (\tilde{\mathbf{x}}, \Omega) 
\longmapsto (p, \rho, u, v) 
\end{align}
where the definition of $\tilde{\mathbf{x}}$ and $\Omega$ follows Eq. \eqref{eq_ArGEnT_trunk}.

\begin{figure}[htbp]
    \centering
    \includegraphics[width=15cm, trim=0cm 0cm 0cm 0cm, clip=true]{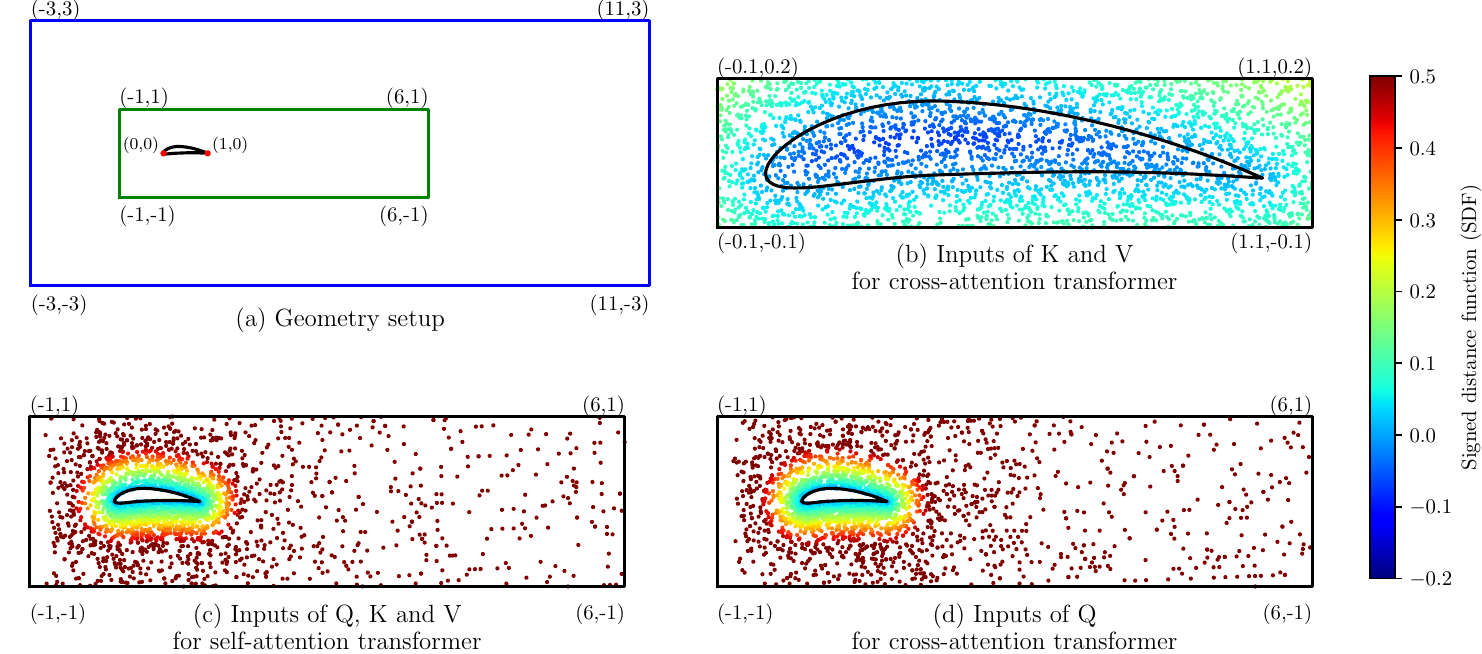}
    \caption{Laminar airfoil flow: (a) Geometry setup. (b, d) Inputs to the cross-attention ArGEnT. (c) Inputs to the self-attention ArGEnT. In (a), the blue box marks the computational domain for numerical simulations, while the green box denotes the region of interest used for training and evaluation. In (b–d), the point coordinates and their associated SDF values serve as inputs to the ArGEnT models. Note that in (b), the geometry points for the keys and values (K and V) can be sampled independently of the query points in (d), using only the point cloud near the airfoil to represent the geometry.}
    \label{Fig_airfoil_setup}
    
\end{figure}

Following \cite{SHUKLA2024107615}, we use 40 samples for training and 10 for testing; the corresponding airfoil profiles are shown in Figure \ref{Fig_airfoil_profiles}(a). An example geometry setup for simulation and training is illustrated in Figure \ref{Fig_airfoil_setup}(a), where the airfoil chord is fixed along the $x$-axis from $(0,0)$ to $(1,0)$. For training and evaluation, we consider only the region $([-1,6]\times[-1,1])$ surrounding the airfoil in order to reduce computational cost. Even so, each case still contains approximately 100,000–150,000 data points. Because the CFD simulations use non-dimensional governing equations, both the spatial coordinates and the flow variables are of $\mathcal{O}(1)$, and no additional normalization is applied. Since this problem involves geometric variation only, we use ArGEnT without a branch network. The four flow variables—pressure $p$, density $\rho$, and velocity components $u$ and $v$—are trained separately using independent model instances to avoid interference among variables.

For the self-attention ArGEnT model, the input consists of the spatial coordinates together with their corresponding SDF values, as illustrated in Fig. \ref{Fig_airfoil_setup}(c). In contrast, the cross-attention and hybrid-attention ArGEnT models adopt a decoupled sampling strategy for building the attention matrices. Specifically, 3000 geometry points are randomly sampled in the vicinity of the airfoil surface, since geometric variations are primarily localized in this region. The query points are sampled independently over the entire training domain, as illustrated in Fig. \ref{Fig_airfoil_setup}(b) and (d).

\begin{figure}[htbp]
    \centering
    \includegraphics[width=15cm, trim=3cm 2.5cm 3cm 3cm, clip=true]{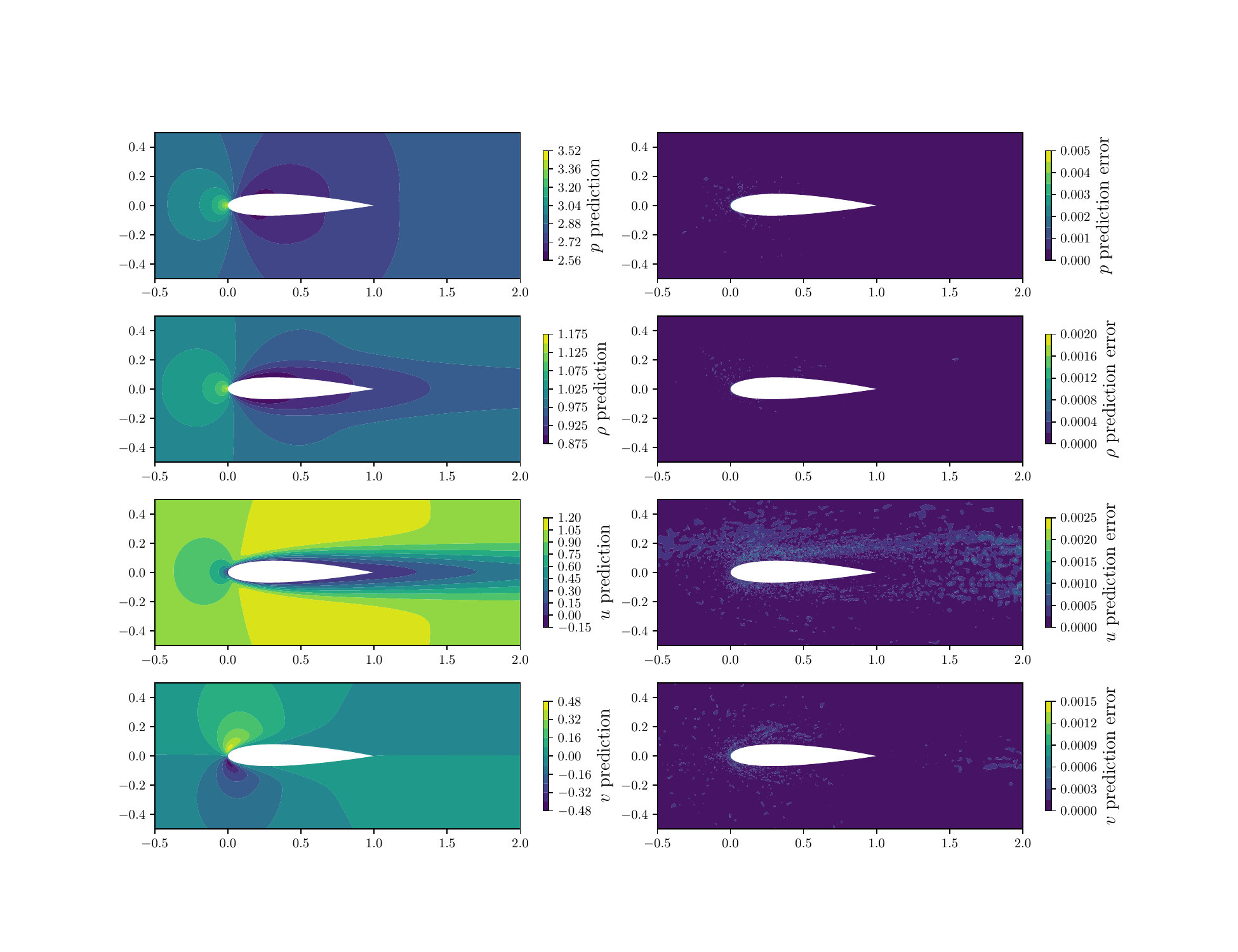}
    \caption{Laminar airfoil flow: contour plots of predicted flow fields (left panels) and predicted absolute errors for a test case using the cross-attention transformer model. Note that all flow field variables are presented in non-dimensional form.}
    \label{Fig_casePrediction}    
\end{figure}

The prediction errors on the test set are summarized in Table \ref{tab_airfoil_errors}, which also includes the results reported in \cite{SHUKLA2024107615} for comparison. The NURBS-DeepONet and Parameter-DeepONet in \cite{SHUKLA2024107615} are two DeepONet variants that use different geometry representations as inputs to the branch network: the NURBS-DeepONet uses 30 ordered airfoil surface points, whereas the Parameter-DeepONet uses the two shape parameters, namely maximum camber and the location of maximum camber. Table \ref{tab_airfoil_errors} shows that all ArGEnT variants significantly outperform these DeepONet baselines, with cross-attention ArGEnT achieving the best accuracy overall. Even the self-attention ArGEnT performs substantially better than the DeepONet variants. Figure \ref{Fig_casePrediction} presents the predicted flow fields and corresponding absolute errors for a representative test case using the cross-attention ArGEnT model, showing that the model captures the main flow features accurately with low errors across the domain.

\begin{figure}[htbp]
    \centering
    \includegraphics[width=12cm, trim=1.0cm 0cm 3cm 1cm, clip=true]{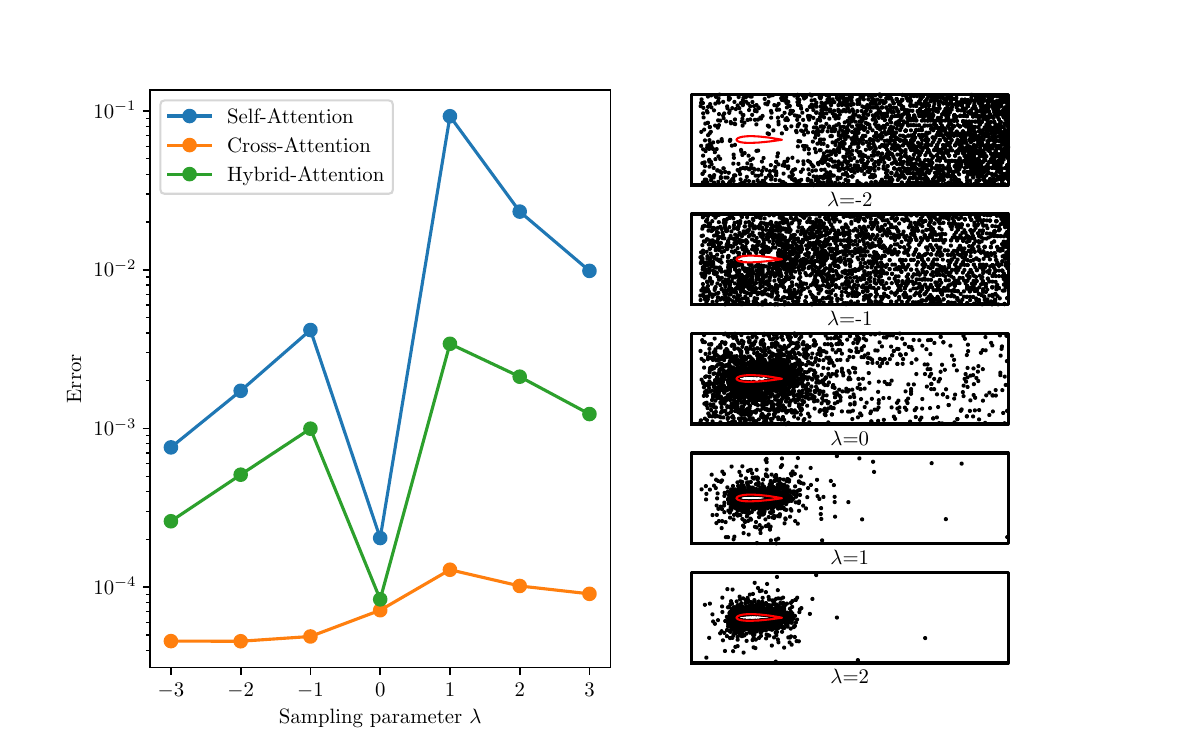}
    \caption{Laminar airfoil flow: effect of sampling strategies on evaluation accuracy of the pressure field. The left panel shows the relative $L_2$ error versus the sampling parameter $\lambda$. The right panels show the corresponding point distributions, where points are sampled using
    $P \propto \frac{1}{1 + 100\, \max(SDF, 10^{-8})^\lambda}$,
    with $SDF$ the signed distance function (negative inside the airfoil, positive outside). When $\lambda=0$, the distribution follows the simulation grid and clusters near the surface; $\lambda>0$ increases clustering, while $\lambda<0$ reduces the clustering near the airfoil and spreads points more into the far field.}
    \label{Fig_accuracy_vs_sampling}    
\end{figure}

In the self- or hybrid-attention ArGEnT models, the point distribution of the query points provides implicit geometric information that influences the attention scores and, consequently, the model's predictions. Although both models achieve good predictive accuracy, the self-attention mechanism in them requires the query points at the inference stage to be sampled from the same distribution as the training data, which limits its flexibility in practical applications since the training data are typically generated from CFD simulations where the query points are determined by a meshing tool used to discretize the computational domain. In contrast, the cross-attention ArGEnT model allows the query points to be sampled independently of the geometry representation, providing substantially greater flexibility in selecting evaluation points. To further investigate the impact of point sampling strategies on evaluation accuracy, we conduct a systematic study by varying the sampling distribution of the query points using a parametrized sampling strategy based on the signed distance function (SDF) to control the degree of point clustering near the airfoil surface. The relative $L_2$ error and the corresponding point distributions as functions of the sampling parameter $\lambda$ are shown in Figure \ref{Fig_accuracy_vs_sampling}. The results indicate that the evaluation accuracy of the self-attention ArGEnT model is highly sensitive to the point sampling distribution, with the lowest error achieved when the query points follow the same distribution as the training data ($\lambda = 0$). Increasing the clustering of points near the airfoil surface ($\lambda > 0$) leads to higher errors, likely due to insufficient coverage of the far-field region, while distributing points more uniformly throughout the domain ($\lambda < 0$) also degrades accuracy because fewer points are allocated to the near-airfoil region where the flow exhibits more complex features. In contrast, the cross-attention transformer is considerably less sensitive to the sampling strategy and maintains higher accuracy across a wide range of sampling distributions, further demonstrating its advantage in handling arbitrary query points independently of the geometry representation.

\begin{table}[h]
\centering
\caption{Laminar airfoil flow: relative $L^2$ errors of field variables for different models on the test set. \enquote{Self-attention}, \enquote{Cross-attention}, and \enquote{Hybrid-attention} refer to ArGEnT with different attention mechanisms.}
\begin{tabular}{ccccc}
\hline
Model
& $p \; (\times 10^{-3})$ 
& $\rho \; (\times 10^{-3})$ 
& $u \; (\times 10^{-3})$ 
& $v \; (\times 10^{-3})$ \\
\hline
NURBS-DeepONet \cite{SHUKLA2024107615}      & 6.05 & 5.89 & 6.21 & 4.60 \\
Parameter-DeepONet \cite{SHUKLA2024107615} & 6.85 & 5.18 & 5.38 & 4.25 \\
Self-attention                  & 0.35 & 0.26 & 1.46 & 4.27 \\
Cross-attention                 & \textbf{0.13} & \textbf{0.11} & \textbf{0.44} & \textbf{1.68} \\
Hybrid-attention                & 0.16 & 0.15 & 0.67 & 2.17 \\
\hline
\end{tabular}
\label{tab_airfoil_errors}
\end{table}

\subsection{Turbulent flow over airfoil of varying shapes and freestream velocities}\label{sec_turbulent_airfoil}
We next consider turbulent flow over airfoils with varying profiles and freestream velocities in order to evaluate the ability of ArGEnT to generalize across both geometric and non-geometric variations. We use the \enquote{full data regime} dataset provided by \cite{bonnet2022airfrans}, which contains 1,000 airfoil profiles, shown in Fig. \ref{Fig_airfoil_profiles}(b), generated from the National Advisory Committee for Aeronautics (NACA) 4-digit and 5-digit series. The freestream velocity varies in both magnitude and angle of attack, with Reynolds number ranging from $2\times10^{6}$ to $6\times10^{6}$ and angle of attack spanning from $-5^\circ$ to $15^\circ$. The Mach number is kept below 0.3 to maintain the incompressible-flow assumption. The flow fields are simulated using a high-fidelity CFD solver with the $k$--$\omega$ SST turbulence model, and the computational domain extends up to 200 chord lengths from the airfoil. The resulting Reynolds-averaged velocity components $(\bar{u}, \bar{v})$, pressure $\bar{p}$, and turbulent viscosity $\nu_t$ serve as the reference solutions for training and evaluation. Further details on dataset generation and simulation setup are given in~\cite{bonnet2022airfrans}.

The learned operator for the following turbulent airfoil problem, taking the cross-attention ArGEnT as an example,  is therefore defined as
\begin{align}
\textbf{Cross-attention} \quad \mathcal{G} : (\tilde{\mathbf{x}}, \Omega, \pmb{\mu}) 
\longmapsto (\bar{u},\, \bar{v},\, \bar{p},\, \nu_t) 
\end{align}
where $\pmb{\mu} =(U_{\inf}, V_{\inf})$ is the far-field freestream velocity. The definition of $\tilde{\mathbf{x}}$ and $\Omega$ follows Eqs. \eqref{eq_ArGEnT_trunk} and \eqref{eq_ArGEnT}.

\begin{figure}[htbp]
    \centering
    \includegraphics[width=15cm, trim=0cm 0cm 0cm 0cm, clip=true]{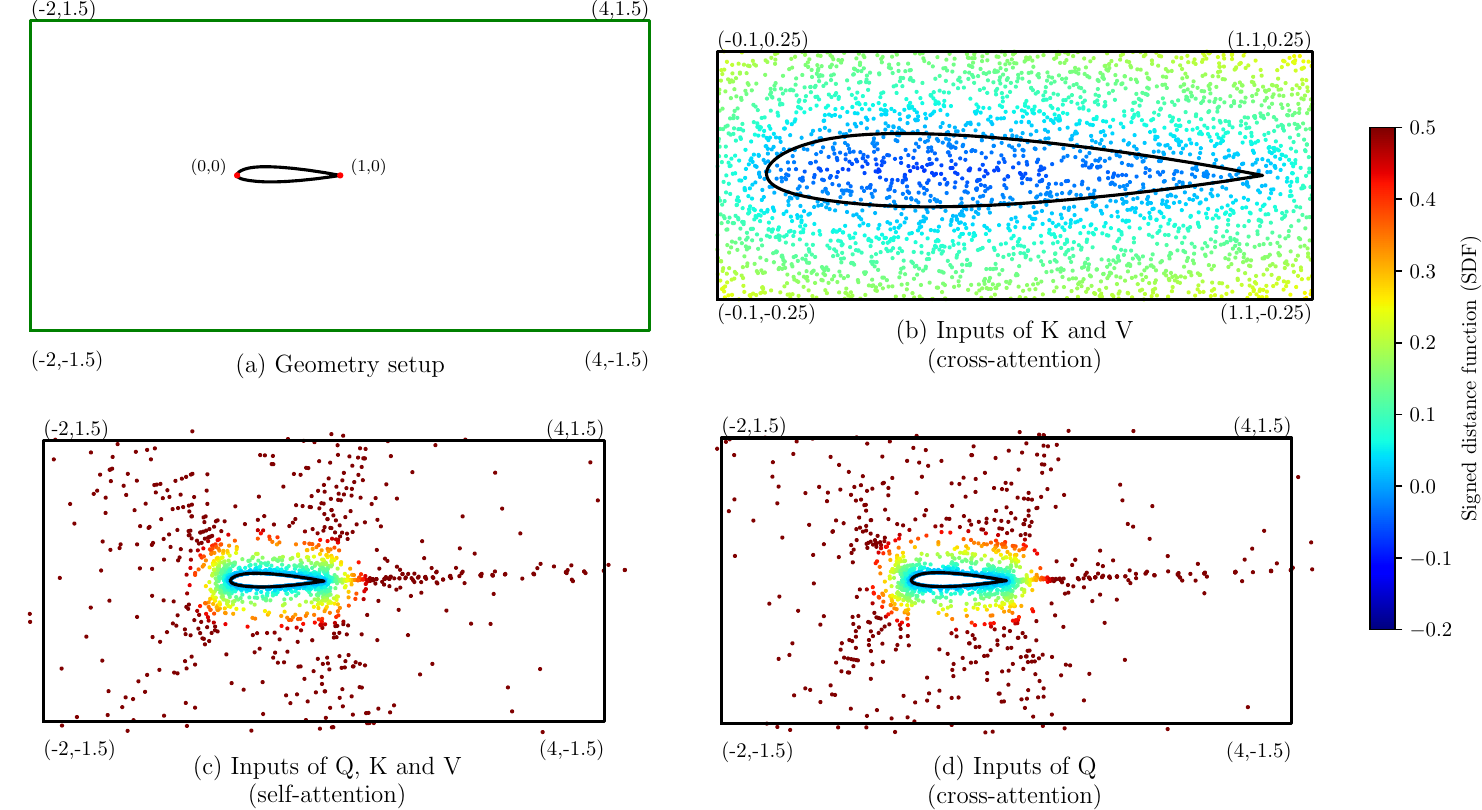}
    \caption{Turbulent airfoil flow: (a) Geometry setup. (b, d) Inputs to the cross-attention transformer. (c) Inputs to the self-attention transformer. In (a), the green box denotes the region of interest used for training and evaluation. In (b–d), the point coordinates and their associated SDF values serve as inputs to ArGEnT models. Note that in (b), the geometry points for the keys and values (K and V) can be sampled independently of the query points in (d), using only the point cloud near the airfoil to represent the geometry.}
    \label{Fig_turbulent_airfoil_setup}
    
\end{figure}

Following the setup in \cite{bonnet2022airfrans}, the \enquote{full data regime} dataset comprises 800 cases for training and 200 cases for testing. The region of interest for both training and evaluation is defined as $[-2,4]\times[-1.5,1.5]$ around the airfoil. The airfoil chord is fixed along the $x$-axis from $(0,0)$ to $(1,0)$, as illustrated in Fig. \ref{Fig_turbulent_airfoil_setup}(a). All solution fields are normalized using $z$-score normalization (as defined in \ref{sec_norm}) for training and evaluation. For the self-attention ArGEnT model, the input consists of the spatial coordinates together with their corresponding SDF values, as shown in Figure~\ref{Fig_turbulent_airfoil_setup}(c). In contrast, for the cross-attention and hybrid-attention ArGEnT models, 3000 geometry points are randomly sampled in the vicinity of the airfoil surface to construct the key and value matrices in the attention blocks, since geometric variations are primarily localized in this region. The query points are sampled independently over the entire training domain, as illustrated in Fig. \ref{Fig_turbulent_airfoil_setup}(b) and~(d).

\begin{figure}[htbp]
    \centering
    \includegraphics[width=15cm, trim=3cm 2.5cm 3cm 3cm, clip=true]{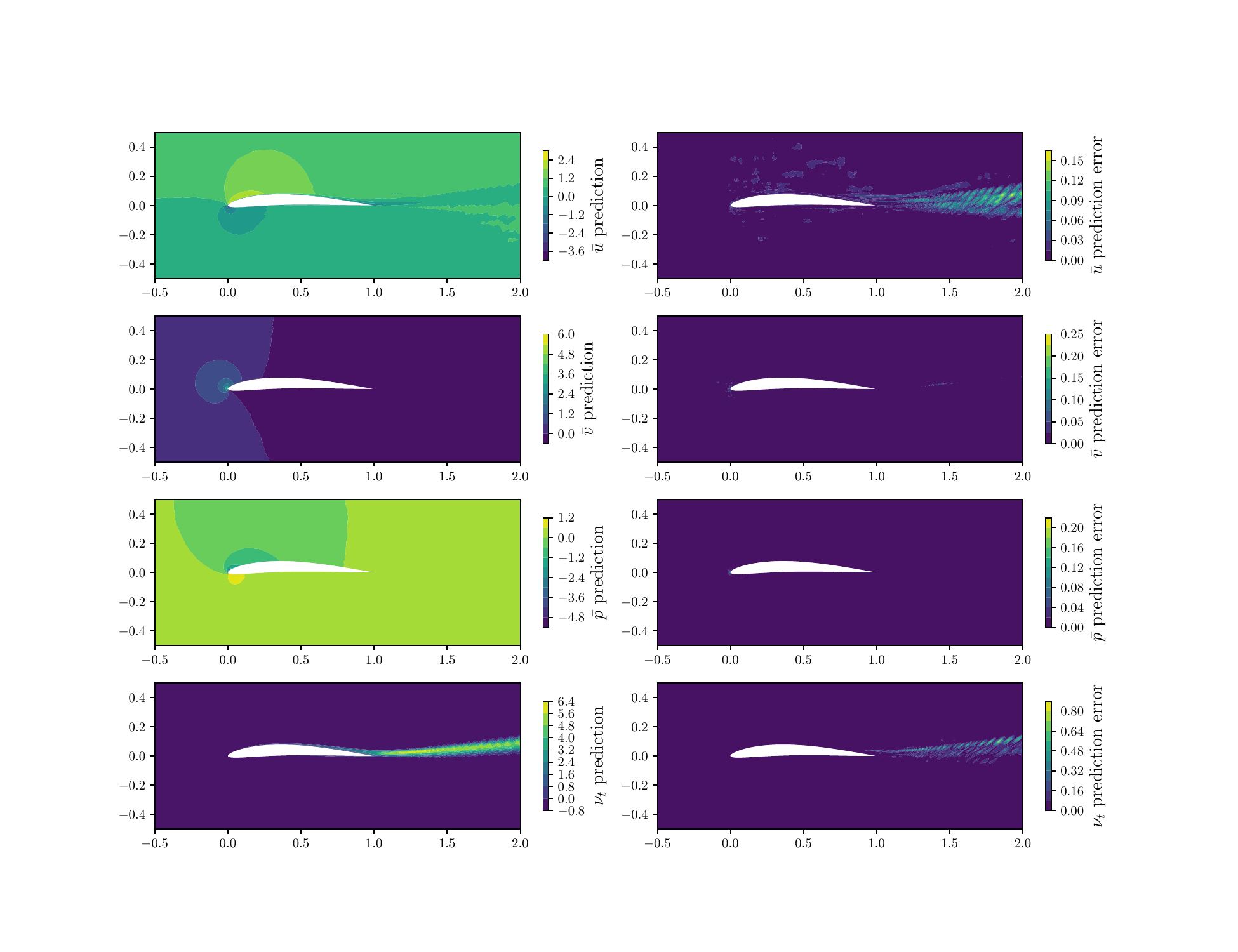}
    \caption{Turbulent airfoil flow: contour plots of predicted flow fields (left panels) and predicted absolute errors for a test case using the cross-attention ArGEnT DeepONet. Note that all flow field variables are presented in normalized form.}
    \label{Fig_turbulent_airfoil_casePrediction}    
\end{figure}

The prediction errors on the test set are summarized in Table~\ref{tab_turbulent_airfoil_errors}, which also includes results from Transolver\cite{wu2024transolver}/Transolver+\cite{luo2025transolver++}. Overall, ArGEnT DeepONet consistently outperforms other widely used baselines, achieving 2--100$\times$ reductions in prediction error across the test cases. The only exception is that Transolver attains a lower error for the horizontal velocity component $\bar{u}$; however, its performance is substantially worse than ArGEnT-DeepONet for the other three variables. Transolver and Transolver+ also require significantly greater computational resources: they are trained for 320,000 steps with a batch size of 1, contain 6.0 million and 3.3 million trainable parameters, respectively, and require 24.8 and 27.5 hours of training. By comparison, ArGEnT-DeepONet is trained for 100,000 steps with a batch size of 3, contains 0.87 million parameters, and requires only about 7.2 hours of training. Among the three attention mechanisms, the lowest errors are obtained with either the cross-attention or hybrid-attention models, while the self-attention model yields comparable accuracy. Figure~\ref{Fig_turbulent_airfoil_casePrediction} presents the predicted flow fields and corresponding absolute errors for a representative test case using cross-attention ArGEnT. The larger errors observed in the wake region are likely caused by the stronger turbulence and flow separation in that area, which typically require either higher-resolution data or more sophisticated modeling to resolve accurately. Overall, these results demonstrate the effectiveness of ArGEnT for predicting turbulent flows over airfoils with varying geometries and freestream conditions.

\begin{figure}[htbp]
    \centering
    \includegraphics[width=12cm, trim=1.0cm 0cm 3cm 1cm, clip=true]{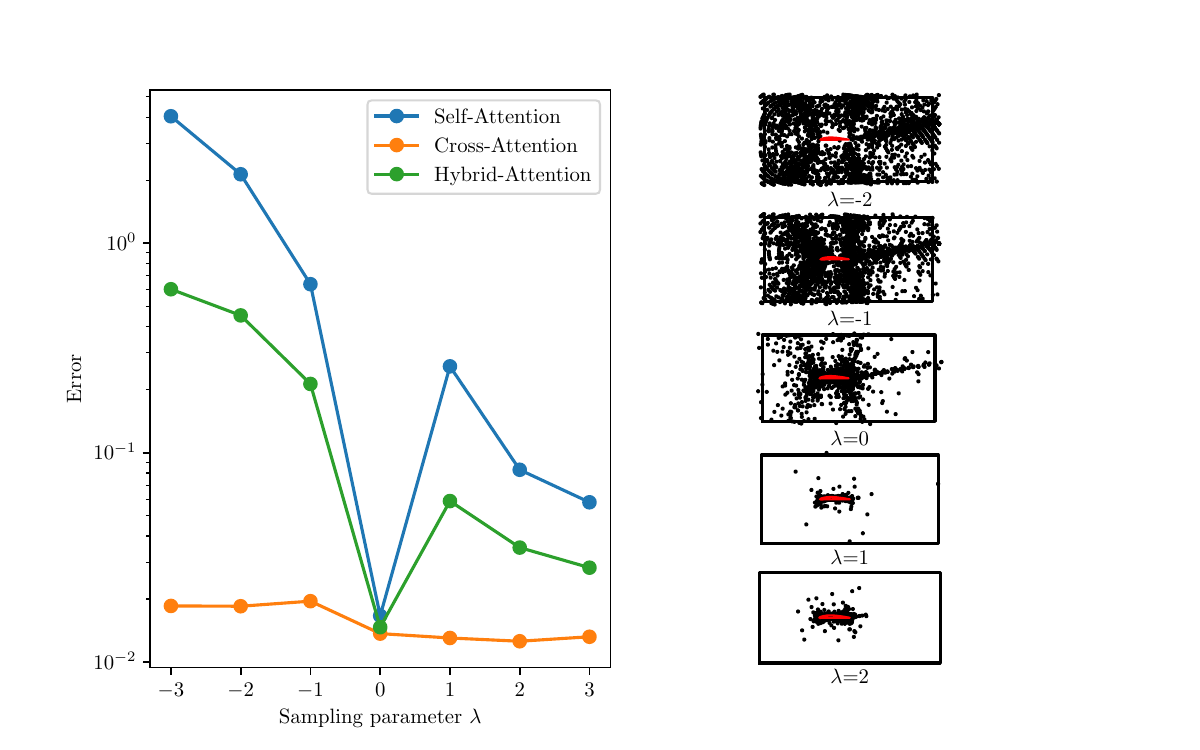}
    \caption{Turbulent airfoil flow: effect of sampling strategies on evaluation accuracy of the pressure field. The left panel shows the relative $L_2$ error versus the sampling parameter $\lambda$. The right panels show the corresponding point distributions, where points are sampled using
    $P \propto \frac{1}{1 + 100\, \max(SDF, 10^{-8})^\lambda}$,
    with $SDF$ the signed distance function (negative inside the airfoil, positive outside). When $\lambda=0$, the distribution follows the simulation grid and clusters near the surface; $\lambda>0$ increases clustering, while $\lambda<0$ reduces the clustering near the airfoil and spreads points more into the far field.}
    \label{Fig_turbulent_airfoil_accuracy_vs_sampling}    
\end{figure}

As in the laminar airfoil study, we further examine how point-sampling strategies affect evaluation accuracy by systematically varying the sampling distribution of query points. Specifically, we adopt a parameterized sampling strategy based on the signed distance function (SDF) to control the clustering of points near the airfoil surface. Figure~\ref{Fig_turbulent_airfoil_accuracy_vs_sampling} shows the relative $L_2$ error together with the corresponding point distributions as a function of the sampling parameter $\lambda$.
The results indicate that the evaluation accuracy of the self-attention and hybrid-attention ArGEnT models is sensitive to the query-point distribution, with the best performance achieved when the sampling distribution matches that used during training ($\lambda=0$). Increasing the clustering of points near the airfoil surface ($\lambda>0$) leads to higher errors, likely because the far field is covered less adequately. Conversely, distributing points more uniformly throughout the domain ($\lambda<0$) also degrades accuracy because fewer points remain concentrated near the airfoil, where the flow features are most complex.
These results highlight the importance of point sampling for achieving optimal accuracy in self-attention and hybrid-attention models. In contrast, cross-attention ArGEnT is substantially less sensitive to the sampling strategy and maintains superior accuracy across a wide range of query-point distributions. This further demonstrates the advantage of cross-attention for enabling accurate predictions at arbitrary query points independently of the geometry representation.

\begin{table}[h]
\centering
\caption{Turbulent airfoil flow: Mean squared errors of normalized field variables for different models on the test set. \enquote{Self-attention}, \enquote{Cross-attention}, and \enquote{Hybrid-attention} refer to the ArGEnT DeepONet with different attention mechanisms.}
\begin{tabular}{ccccc}
\hline
Model & $\bar{u} (\times 10^{-2})$
      & $\bar{v} (\times 10^{-2})$
      & $\bar{p} (\times 10^{-2})$
      & $\nu_t   (\times 10^{-2})$ \\
\hline
MLP\cite{bonnet2022airfrans}          & 0.95  & 0.98 & 0.74 & 1.90 \\
GraphSAGE\cite{bonnet2022airfrans}     & 0.83 & 0.99 & 0.66 & 1.60 \\
PointNet\cite{bonnet2022airfrans}      & 3.50 & 3.64 & 1.15 & 2.92 \\
Graph U-Net\cite{bonnet2022airfrans}   & 1.52 & 2.03 & 0.66 & 1.46 \\
Transolver & \textbf{0.01}	& 12.10 &	28.10	&7.18 \\
Transolver+ & 0.08	& 0.44 	&   22.4	&0.81 \\
Self-attention          & 0.031 & 0.027 & 0.229 & 0.717 \\
Cross-attention         & 0.027 & 0.027 & \textbf{0.029} & 0.640 \\
Hybrid-attention        & 0.076 & \textbf{0.027} & 0.081 & \textbf{0.440} \\
\hline
\end{tabular}
\label{tab_turbulent_airfoil_errors}
\end{table}

\subsection{Lid-driven cavity flow}\label{sec_lid}
\begin{figure}[htbp]
    \centering
    \includegraphics[width=15cm, trim=0cm 0cm 0cm 0cm, clip=true]{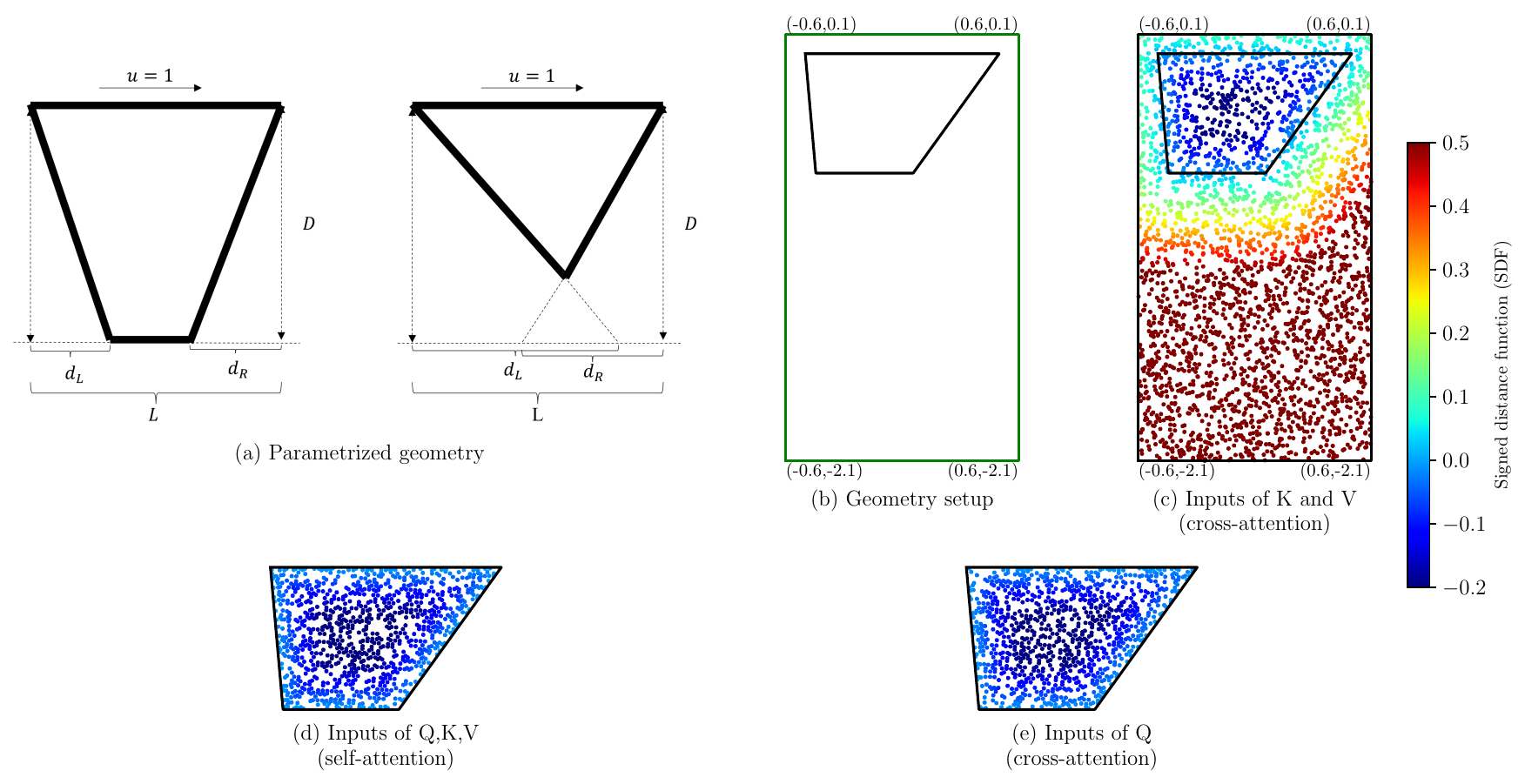}
    \caption{Lid driven flow: (a) Parametrized geometry. (b) Geometry setup. (c, e) Inputs to the cross-attention transformer. (d) Inputs to the self-attention transformer. In (b) the green box denotes the region where geometry points are sampled. In (c-e), the point coordinates and their associated SDF values serve as inputs to ArGEnT models. Note that in (c), the geometry points for the keys and values (K and V) can be sampled independently of the query points in (e), using only the point spreading over all the possible computational domains.}
    \label{Fig_lid_setup}    
\end{figure}

Lid-driven cavity flow is a classic benchmark in fluid dynamics, consisting of a closed cavity in which the flow is driven by the motion of the top lid~\cite{reddy2022finite}. Here we consider a two-dimensional cavity with parameterized geometries and varying Reynolds numbers. As shown in Fig.~\ref{Fig_lid_setup}(a), the cavity geometry is defined on an $L \times D$ rectangular domain trimmed by two triangular cutouts at the bottom corners, parameterized by $d_L$ and $d_R$. The flow is driven by the top lid, which moves at the constant non-dimensional velocity $(u,v)=(1,0)$.
The lid width is fixed at $L=1$, while the remaining geometric parameters $(D, d_L, d_R) \in [0.5,2] \times [0,1] \times [0,1]$ are varied to generate a family of cavity shapes. When $d_L + d_R < 1$, the resulting domain has a trapezoidal shape, whereas for $d_L + d_R \geq 1$, the domain becomes triangular. The Reynolds number is varied in the range $100 \leq \mathrm{Re} \leq 1000$.
The flow fields are simulated using OpenFOAM\textsuperscript{\textregistered}, solving the non-dimensional incompressible Navier–Stokes equations. The resulting velocity components $(u,v)$ and pressure field $p$ provide the high-fidelity solutions used to train the models and assess their predictive performance.

The learned operator for the following lid-driven cavity flow problem, taking the cross-attention ArGEnT as an example,  is therefore defined as
\begin{align}
\textbf{Cross-attention} \quad \mathcal{G} : (\tilde{\mathbf{x}}, \Omega, \pmb{\mu}) 
\longmapsto (u, v, p) 
\end{align}
where $\pmb{\mu} =(Re,)$ is the Reynolds number. The definition of $\tilde{\mathbf{x}}$ and $\Omega$ follows Eqs. \eqref{eq_ArGEnT_trunk} and \eqref{eq_ArGEnT}.

We randomly generate 3,000 cavity geometries and Reynolds numbers, using 80\% of the samples for training and the remaining 20\% for testing. Because the CFD simulations employ non-dimensional governing equations, the spatial coordinates and flow variables are both of order 1, so no additional normalization is applied. An example of the ArGEnT input setup is shown in Figure \ref{Fig_lid_setup}(b--e). For the cross- and hybrid-attention ArGEnT models, the geometry points are sampled over a larger region to cover all possible cavity shapes. The query points are sampled inside the cavity domain independently of the geometry points, as illustrated in Figures \ref{Fig_lid_setup}(c) and (e). 

The prediction errors on the test set are summarized in Table~\ref{tab_lid_errors}, where results from the standard DeepONet, Point-DeepONet~\cite{park2026point}, and Transolver\cite{wu2024transolver}/Transolver+\cite{luo2025transolver++} are included for comparison. All the models are configured to have comparable parameter counts (0.46--0.87 million) and are trained for 100,000 steps (approximately 1,333 epochs) with a batch size of 32. Transolver and Transolver+ follow their original configurations, using a batch size of 1 and 960,000 training steps (400 epochs).

The results show that Point-DeepONet yields even higher prediction errors than the standard DeepONet. One possible explanation is that, when a suitable geometric parameterization is available, representing geometry through explicit geometric parameters can be more accurate than using point-cloud representations. Nevertheless, Point-DeepONet retains significant flexibility for handling arbitrary geometries. Transolver and Transolver+ achieve accuracy comparable to that of the ArGEnT models, but at substantially higher computational cost because of their much larger model sizes and longer training schedules. All ArGEnT models consistently outperform both the standard DeepONet and Point-DeepONet, achieving substantially lower prediction errors. Among the attention-based variants, the cross-attention and hybrid-attention models deliver the best accuracy, while the self-attention model remains competitive. Relative to Point-DeepONet, the cross- and hybrid-attention ArGEnT models use a fixed point cloud augmented with signed distance function (SDF) values, which provides a more accurate and consistent geometric representation. In terms of computational cost, the standard DeepONet remains the most efficient because of its simpler architecture.

\begin{figure}[htbp]
    \centering
    \includegraphics[width=15cm, trim=3cm 1.5cm 1cm 1.5cm, clip=true]{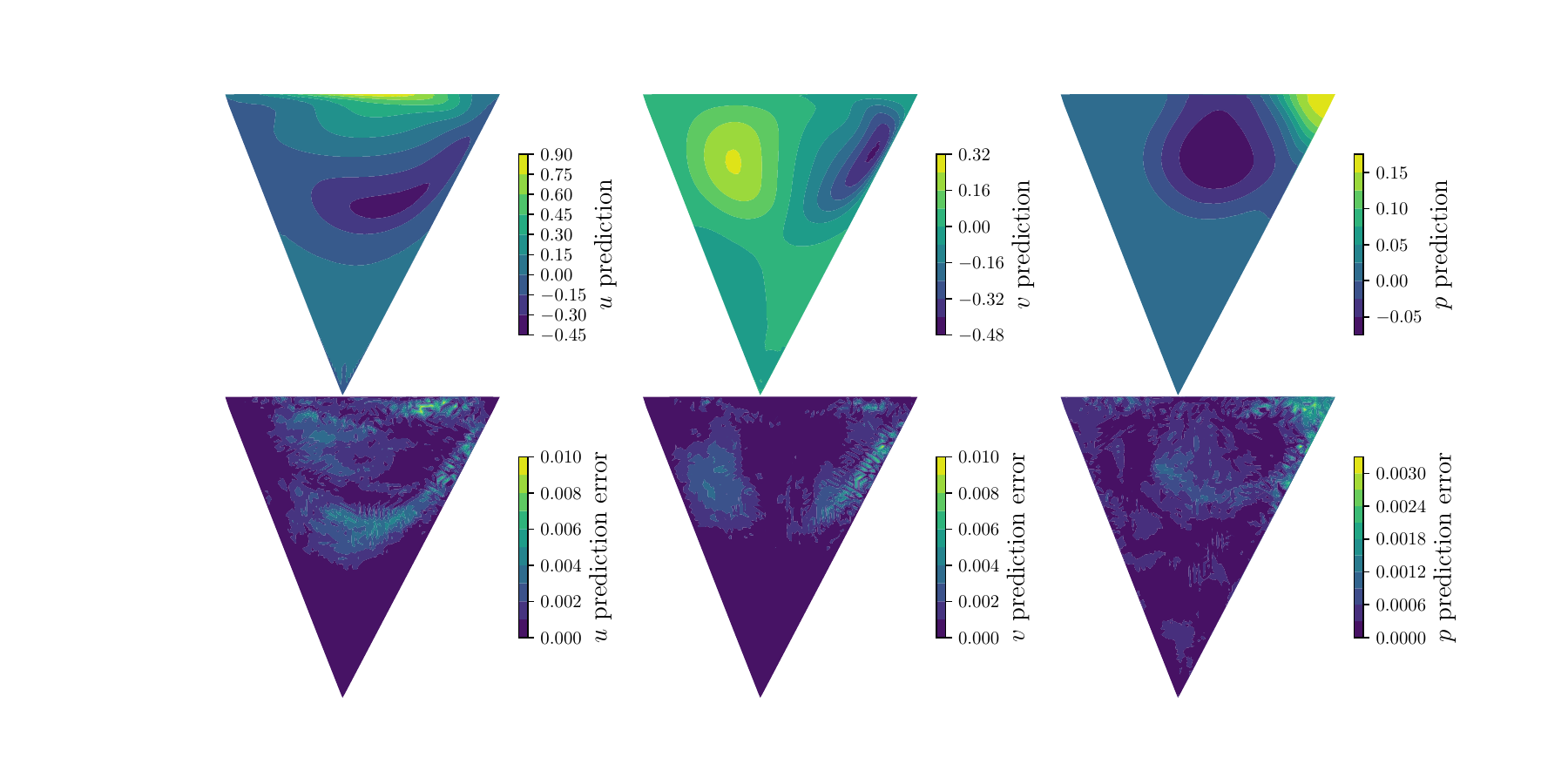}
    \caption{Lid driven flow: contour plots of predicted flow fields (upper panels) and predicted absolute errors (bottom panels) for a test case using the cross-attention transformer model. Note that all flow field variables are presented in non-dimensional form.}
    \label{Fig_lid_casePrediction}    
\end{figure}

Figure~\ref{Fig_lid_casePrediction} presents the predicted flow fields and the corresponding absolute error distributions for a representative test case obtained using the cross-attention ArGEnT DeepONet. The model accurately captures the vortex structures within the cavity, while the largest errors are primarily localized near the upper-right corner, where strong velocity gradients arise from the interaction between the moving lid and the stationary wall.

\begin{figure}[htbp]
    \centering
    \includegraphics[width=15cm, trim=3cm 2cm 1cm 1.5cm, clip=true]{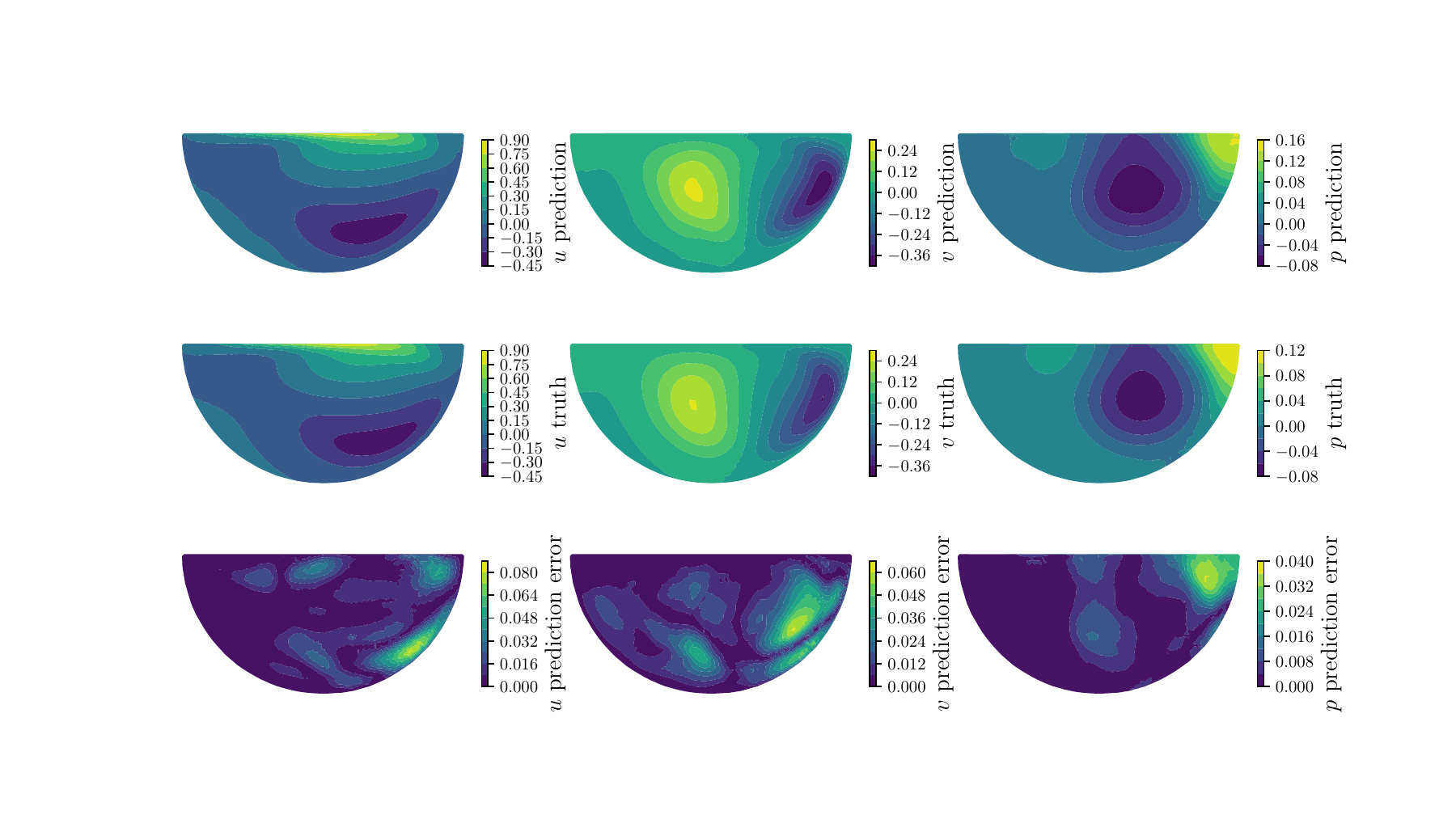}
    \caption{Lid driven flow: contour plots of predicted flow fields (upper panels), reference (middle panels) and predicted absolute errors (bottom panels) for a test case using the cross-attention transformer model. The test geometry lies outside the expressive capacity of the employed geometry parameterization  for training, though its Reynolds number (924) remains within the training range. Note that all flow field variables are presented in non-dimensional form.}
    \label{Fig_lid_casePrediction_Extrapolation}    
\end{figure}

To further assess the generalization capability of the ArGEnT models, we evaluate the trained models on a case that lies outside the expressive capacity of the geometry parameterization used in training, while keeping the Reynolds number (924) within the training range. The results are shown in Figure \ref{Fig_lid_casePrediction_Extrapolation}. Cross-attention ArGEnT still provides reasonable predictions, although with larger errors than those observed for test cases within the training geometry distribution. This demonstrates the potential of ArGEnT to generalize to unseen geometries. The self- and hybrid-attention models, not shown here, exhibit similar behavior, with slightly higher errors than the cross-attention model in this extrapolation case. By contrast, the standard DeepONet cannot be applied to this case because it relies on a predefined geometry parameterization, and the extrapolated geometry cannot be represented within that parameterization. This limitation highlights a key advantage of the attention-based models: they operate directly on more general geometric representations and therefore offer greater flexibility in handling arbitrary geometries. This advantage is particularly relevant in practical applications, where geometric variation can be substantial and difficult to capture using a fixed low-dimensional parameterization. Moreover, these attention-based models could potentially be strengthened further through transfer learning \cite{niu2021decade,goswami2022deep} or continual learning \cite{parisi2019continual, hadsell2020embracing, fu2024physics}, which may improve their ability to adapt to unseen geometries or related problems.

\begin{table}[htbp]
\centering
\caption{Lid driven flow: relative $L^2$ errors of field variables for different models on the test set. \enquote{Self-attention}, \enquote{Cross-attention}, and \enquote{Hybrid-attention} refer to the ArGEnT DeepONet with different attention mechanisms, while \enquote{DeepONet} refers to the standard DeepONet model.}
\label{tab:ablation_updated}
\begin{tabular}{ccccc}
\hline
Model & $u\;(\times 10^{-2})$ & $v\;(\times 10^{-2})$ & $p\;(\times 10^{-2})$ & Time (hours) \\
\hline
DeepONet          & 1.00 & 1.13 & 1.84 & \textbf{0.96}\\
Point-DeepONet & 1.81 & 1.19 & 2.97 & 1.25 \\
Transolver & 0.74 & 0.83 & 2.04 & 23.18 \\
Transolver+ & 0.74 & 0.89 & 2.08 & 23.96 \\
Self-attention   & 0.75 & 0.84 & 1.55 & 4.11 \\
Cross-attention  & \textbf{0.69} & 0.83 & 1.61 & 4.00\\
Hybrid-attention & 0.71 & \textbf{0.79} & \textbf{1.52} & 4.02\\
\hline
\end{tabular}
\label{tab_lid_errors}
\end{table}

\begin{remark}
\textit{Note that while the errors are not significantly different in Table \ref{tab_lid_errors} between the DeepONet and ArGEnT, only ArGEnT can handle geometries without parameterization. This is demonstrated in Figure \ref{Fig_lid_casePrediction_Extrapolation}, {where the hemisphere geometry cannot be represented using the trapezoid/triangle parameterization adopted during training.}
} 
\end{remark}

\subsection{Redox Flow Battery}\label{sec_RFB}
\begin{figure}[htbp]
    \centering
    \includegraphics[width=15cm, trim=0cm 0cm 0cm 0cm, clip=true]{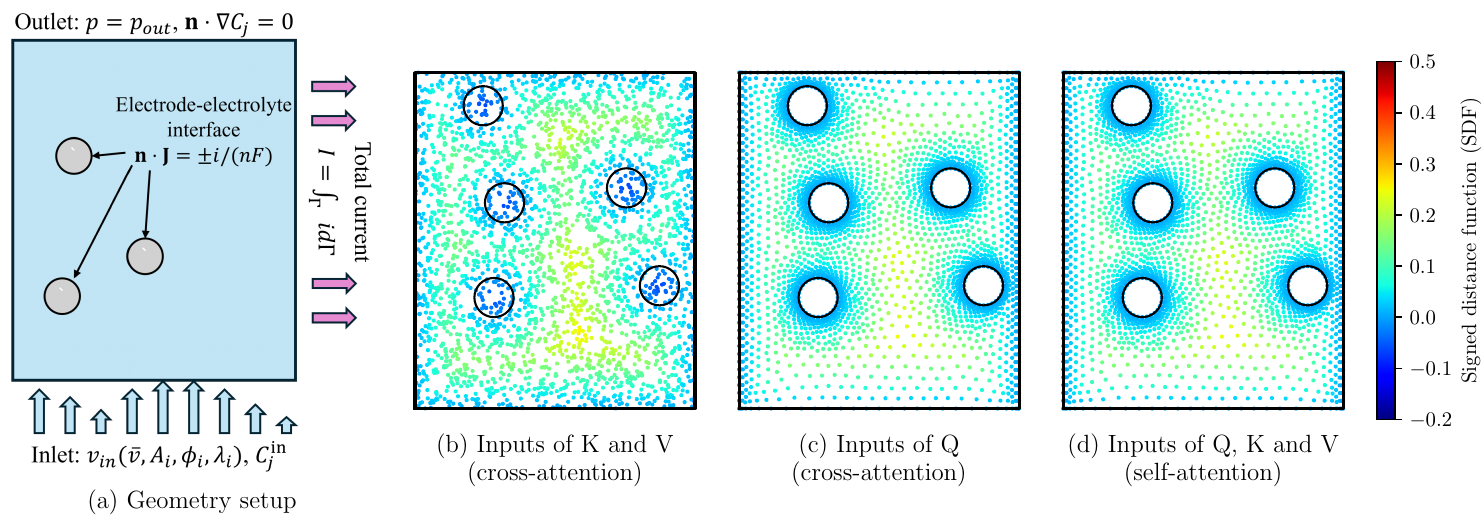}
    \caption{Redox flow battery: (a) Geometry setup and boundary conditions. (b, c) Inputs to the cross-attention transformer. (d) Inputs to the self-attention transformer.  In (b-d), the point coordinates and their associated SDF values serve as inputs to ArGEnT models. Note that in (b), the geometry points for the keys and values (K and V) can be sampled independently of the query points in (c), using the points spreading over all the possible computational domains.}
    \label{Fig_Rods_setup}
    
\end{figure}

Redox flow batteries (RFBs) are rechargeable electrochemical energy-storage systems in which energy is stored in liquid electrolytes containing dissolved redox-active species that are circulated through an electrochemical cell. Electrolyte flow through the porous electrodes plays a critical role in RFB operation because it directly affects mass transport, charge transfer, and overall electrochemical performance. In this study, we consider a simplified two-dimensional RFB model to generate high-fidelity simulation data for testing ArGEnT. The model focuses on the negative half-cell and employs an organic redox species, DHPS~\cite{hollas2018biomimetic,zeng2022characterization}, as the electrochemically active material. The geometry setup is illustrated in Fig. \ref{Fig_Rods_setup}(a). The negative cell consists of a rectangular flow channel with width 100 $\mu\mathrm{m}$ and height 120 $\mu\mathrm{m}$. Several cylindrical rods inside the cell represent the solid phase of the distributed porous electrodes. The rods have a fixed diameter of 14 $\mu\mathrm{m}$, while their number and positions vary from case to case. The electrolyte is pumped into the channel at a nominal inlet velocity of 5 mm/s. A parameterized perturbation is added to the inlet profile, defined as follows:
\begin{equation}
    v_{in} =  \bar{v}(1+A_1\sin(2\pi\lambda_1 x + \phi_1) +A_2\sin(2\pi\lambda_2 x + \phi_2)) 
\end{equation}
where $\bar{v}=5$ mm/s is the mean inlet velocity, $\{A_1, A_2, \phi_1, \phi_2, \lambda_1, \lambda_2\}$ are random parameters controlling the inlet velocity perturbation. The electrochemical reactions will take place at the surface of the rods, consuming the reactant species and generating electric current. The problem is simulated using COMSOL\textsuperscript{\textregistered} that solves the coupled physics comprising fluid flow, species transport, charge transport, and interfacial electrochemical kinetics, as described in \ref{sec_RFB_model}. The resulting velocity ($u$, $v$), pressure ($p$), electric potential ($\phi_e^-$), overpotential ($\eta^-$), and species concentration ($c_R$) fields provide the high-fidelity solutions used to train the models and assess their predictive performance.

The learned operator for the following redox flow battery problem, taking the cross-attention ArGEnT as an example,  is therefore defined as
\begin{align}
\textbf{Cross-attention} \quad \mathcal{G} : (\tilde{\mathbf{x}}, \Omega, \pmb{\mu}) 
\longmapsto (\phi_e^-, \eta^-, c_R, u, v, p) 
\end{align}
where $\pmb{\mu} =\bf{v}_{in}$ is the inlet velocity profile, represented by velocity values sampled at 100 uniformly distributed inlet points. The definition of $\tilde{\mathbf{x}}$ and $\Omega$ follows Eqs. \eqref{eq_ArGEnT_trunk} and \eqref{eq_ArGEnT}.

For $nRods = 1, 3, 5$, we randomly generate 2813, 2568, and 2346 distinct configurations, respectively, each consisting of a geometry and an inlet velocity profile. For each rod count, 80\% of the samples are used for training and the remaining 20\% for testing. Because the solution variables span different magnitudes, $z$-score normalization is applied to each variable, as defined in \ref{sec_norm}. An example of the ArGEnT input setup is shown in Fig. \ref{Fig_Rods_setup}(b--d). For the cross-attention ArGEnT model, geometry points are sampled over the cell to ensure coverage of all possible geometries. The query points are sampled independently within the computational domain, as illustrated in Fig.~\ref{Fig_Rods_setup}(c) and (d). To represent the inlet velocity profile, we uniformly sample 100 points along the inlet boundary and use the corresponding velocity values as inputs to the branch network of the ArGEnT DeepONets.

The prediction errors on the test set are summarized in Table~\ref{tabl_Rods_self_validation_errors}, where results from the standard DeepONet, Point-DeepONet~\cite{park2026point}, and Transolver\cite{wu2024transolver}/Transolver+\cite{luo2025transolver++} are included for comparison. All the models are configured to have comparable parameter counts (0.47--0.87 million) and are trained for 100,000 steps (approximately 1,333 epochs) with a batch size of 32. Transolver and Transolver+ follow their original configurations, using a batch size of 1, and 800,000 training steps (400 epochs).

\begin{figure}[htbp]
    \centering
    \includegraphics[width=15cm, trim=7cm 7cm 6cm 4cm, clip=true]{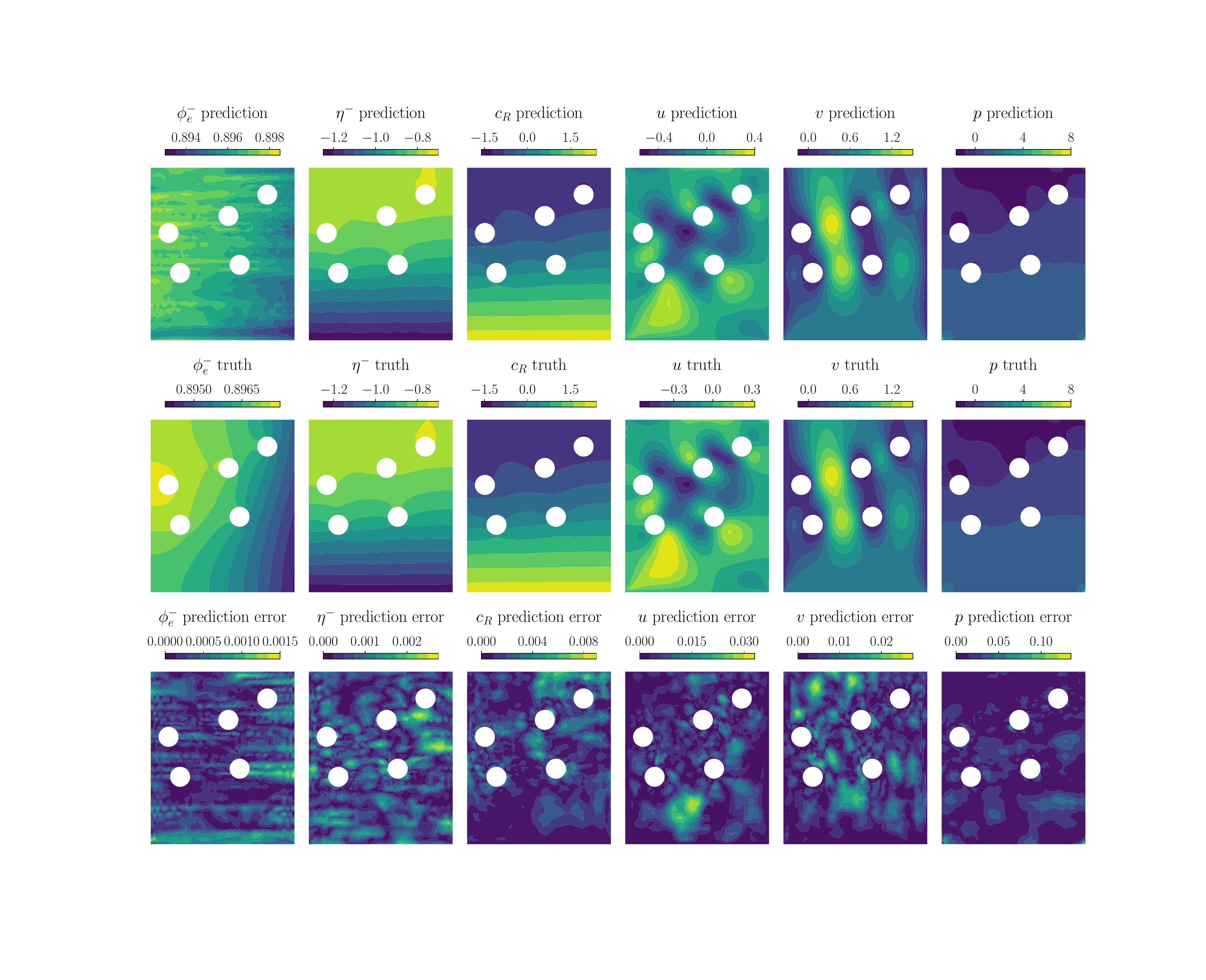}
    \caption{Redox flow battery: contour plots of predicted fields (left panels) and predicted absolute errors for a test case using the cross-attention transformer model. Note that all field variables are presented in normalized form.}
    \label{Fig_Rods_casePrediction}    
\end{figure}

All models are trained separately for each value of $nRods$. For $nRods = 1$, the prediction errors of $\phi_e^-$ obtained with the standard DeepONet are comparable to those of the attention-based models. For the remaining five variables, however, the attention-based models achieve substantially lower errors, in many cases improving by more than one order of magnitude over the standard DeepONet. As the number of rods increases, the prediction errors of all models grow because the geometry becomes more complex. Even so, the attention-based models consistently outperform the standard DeepONet across all variables and all values of $nRods$.
For the flow variables ($u$, $v$, and $p$), the attention-based models achieve approximately 50$\times$ lower errors than the standard DeepONet when $nRods = 3$ and $nRods = 5$, where the standard DeepONet fails to produce reasonable predictions. This behavior can be attributed to the limitation of representing geometry solely through concatenated rod-center coordinates in the DeepONet branch network. In particular, the DeepONet branch network is not permutation invariant with respect to the ordering of rod centers: two geometrically identical configurations specified as $(\mathbf{x}_1,\mathbf{x}_2)$ and $(\mathbf{x}_2,\mathbf{x}_1)$ are treated as distinct inputs despite representing the same physical geometry. As the number of rods increases, this ambiguity leads to inconsistent operator inputs and degraded predictive performance. By contrast, the attention-based models operate directly on unordered geometric representations and are permutation invariant with respect to the ordering of individual geometric elements. This provides a consistent encoding of multi-rod configurations and a clear advantage for handling complex arbitrary geometries.
The standard DeepONet nevertheless yields reasonable predictions for the electrochemical and concentration variables ($\phi_e^-$, $\eta^-$, and $c_R$), as these quantities are less sensitive to rod interactions than the flow field, as illustrated in Figure~\ref{Fig_Rods_casePrediction}. Among the attention-based models, the cross-attention and hybrid-attention variants achieve the highest accuracy, while the self-attention model remains comparable. In terms of computational cost, the standard DeepONet is the most efficient because of its simpler architecture. Similar to the findings in Section~\ref{sec_lid}, Point-DeepONet does not outperform the standard DeepONet, which may be due to the limited geometric fidelity of uniformly sampled point clouds within the computational domain. Transolver and Transolver+ achieve competitive accuracy for some variables, but their performance is inconsistent across the six target fields. In particular, Transolver+ yields much larger pressure errors for $nRods = 1$ and $nRods = 3$ ($1.33\times 10^{-1}$ and $2.10\times 10^{-1}$, respectively), and both Transolver variants remain substantially more expensive computationally. These results indicate that, for the redox flow battery problem, the ArGEnT models provide a more reliable and efficient surrogate overall.

Figure~\ref{Fig_Rods_casePrediction} shows the predicted flow fields and the corresponding absolute error distributions for a representative test case obtained using the cross-attention ArGEnT. The model accurately captures the flow structures around the rods as well as the spatial distributions of the electrochemical variables. An exception is observed for the electrolyte potential $\phi_e^-$, for which the predicted spatial pattern does not fully match the reference solution.
Despite this mismatch in spatial patterns, the absolute prediction errors of $\phi_e^-$ remain trivial. This behavior is  attributed to the fact that $\phi_e^-$ is nearly constant throughout the domain, with spatial variations that are much smaller than its mean value. Such weak variations make it challenging for the model to accurately learn and reproduce the subtle spatial structure of $\phi_e^-$, even though the overall magnitude is well captured.

\begin{figure}[htbp]
    \centering
    \includegraphics[width=15cm, trim=7cm 7cm 6cm 4cm, clip=true]{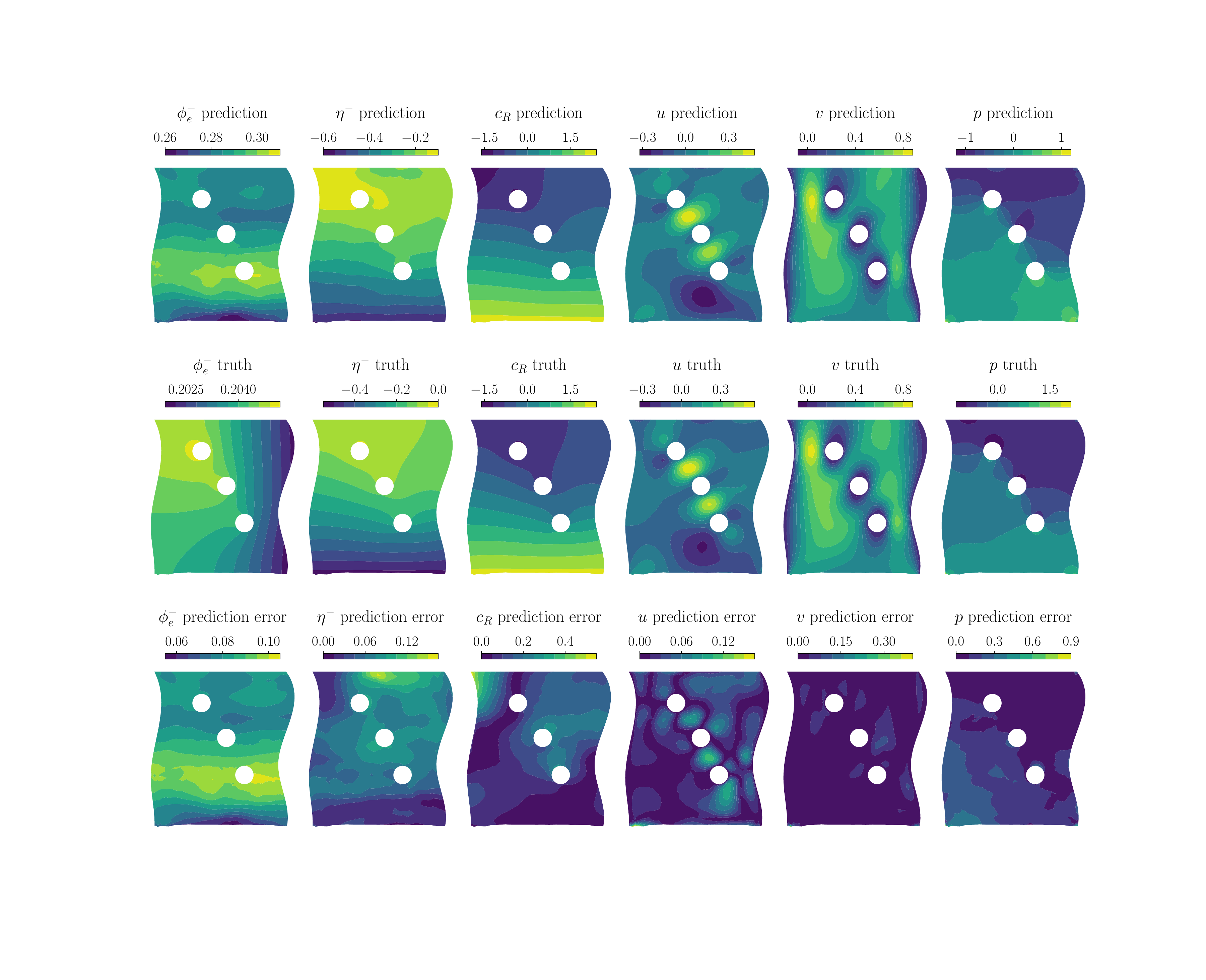}
    \caption{Redox flow battery: contour plots of predicted fields (left panels) and predicted absolute errors for a test case using the cross-attention transformer model. The test geometry lies outside the expressive capacity of the employed geometry parameterization  for training, though its Reynolds number (924) remains within the training range. Note that all field variables are presented in normalized form.}
    \label{Fig_Rods_casePrediction_Extrapolation}    
\end{figure}

To further assess the generalization capability of the ArGEnT models, we evaluate the trained networks on a test case that lies outside the expressive capacity of the geometry parameterization used during training. The results are shown in Figure~\ref{Fig_Rods_casePrediction_Extrapolation}. The cross-attention ArGEnT is still able to produce reasonable predictions when the left and right cell walls become curved, although the prediction errors increase compared to those for test cases within the training geometry distribution. This result demonstrates the potential of the ArGEnT models to generalize to previously unseen geometries.
The self-attention and hybrid-attention models (not shown) exhibit similar qualitative behavior in this extrapolation setting. In contrast, the standard DeepONet fails to handle this extrapolation case, as it relies explicitly on the predefined geometry parameterization employed during training. This limitation highlights a key advantage of ArGEnT models, which can operate directly on more general geometric representations and therefore offer greater flexibility in handling arbitrary geometries.

We further evaluate the extrapolation capability of the ArGEnT models with respect to the number of rods. Separate cross-attention ArGEnT models are trained using datasets with $nRods = 1$, $3$, and $5$, respectively, and each trained model is evaluated on test cases spanning all three rod configurations. The resulting test errors are summarized in Table~\ref{tabl_Rods_cross_validation_errors}.
The results show that models trained on simpler geometries (with fewer rods) do not generalize well to more complex geometries (with a larger number of rods). In contrast, when trained on more complex geometries, the model generalizes reasonably well to simpler configurations. {This asymmetric behavior reflects the different levels of geometric and physical complexity covered by the training distributions. Models trained on one-rod geometries have not encountered multiple solid components, narrow flow passages, or rod–rod interactions and therefore do not generalize well to five-rod configurations. In contrast, five-rod geometries contain a broader range of local geometric and physical patterns, some of which remain relevant when the number of rods is reduced. The five-to-one result should therefore be interpreted as generalization from a richer geometry distribution to a simpler one, rather than as an equally challenging form of extrapolation in the reverse direction.} As expected, the lowest prediction errors are obtained when the training and testing geometries involve the same number of rods.
This asymmetric generalization behavior also extends to extrapolation beyond the training geometry parameterization. In particular, models trained on simpler geometries fail to generalize to the curved-boundary case, whereas those trained on more complex geometries with a larger number of rods achieve reasonable performance. Overall, these results highlight the importance of geometric diversity in the training data for enabling robust extrapolation to unseen geometries.

\begin{landscape}
\begin{table}[p]
\centering
\caption{Redox flow battery problem: relative $L^2$ errors of normalized field variables at different number of rods $nRods$. Best (lowest) values in each column are highlighted in bold. \enquote{Self-attention}, \enquote{Cross-attention}, and \enquote{Hybrid-attention} refer to the ArGEnT DeepONet models with different attention mechanisms, while \enquote{DeepONet} refers to the standard DeepONet model.}
\scriptsize
\setlength{\tabcolsep}{4pt}
\renewcommand{\arraystretch}{0.9}
\begin{tabular}{c c c c c c c c c}
\toprule
$nRods$ & Model 
& $\phi_e^{-}$ & $\eta^{-}$ & $c_R$ & $u$ & $v$ & $p$ & Time (hours) \\
\midrule
\multirow{7}{*}{1}
& DeepONet
& 8.96e-4 & 1.02e-3 & 4.470e-3 & 6.49e-2 & 2.03e-2 & 2.39e-2 & \textbf{0.83} \\
& Point-DeepONet
& 3.52e-1 & 3.72e-1 & 4.09e-1 & 5.17e-2 & 1.06e-1 & 3.77e-1 & {1.11} \\
& Transolver
& 1.55e-3 & 5.14e-4 & 1.95e-3 & 1.79e-2 & 3.90e-3 & 9.67e-3 & 18.47 \\
& Transolver+
& 1.06e-3 & 4.82e-4 & 2.22e-3 & 1.70e-2 & 3.08e-3 & 1.33e-1 & 19.37 \\
& Self-attention
& 6.07e-4 & 5.17e-4 & 2.16e-3 & 1.00e-2 & 3.99e-3 & 6.76e-3 & 3.78 \\
& Cross-attention
& \textbf{3.76e-4} & \textbf{3.23e-4} & \textbf{1.38e-3} & \textbf{7.97e-3} & \textbf{3.05e-3} & \textbf{6.48e-3} & 4.17 \\
& Hybrid-attention
& 1.32e-3 & 5.72e-4 & 2.10e-3 & 1.06e-2 & 3.41e-3 & 6.68e-3 & 3.61 \\
\midrule
\multirow{7}{*}{3}
& DeepONet
& 6.13e-2 & 3.75e-2 & 2.22e-2 & 1.28e0 & 4.63e-1 & 4.76e-1 & \textbf{0.83} \\
& Point-DeepONet
& 1.97e-1 & 3.61e-2 & 3.45e-1 & 1.00e0 & 1.74e-1 & 3.34e-1 & {1.11} \\
& Transolver
& 2.58e-2 & 1.68e-2 & 4.66e-3 & 4.02e-2 & 1.31e-2 & 3.03e-2 & 18.26 \\
& Transolver+
& 2.35e-2 & 1.70e-2 & 5.71e-3 & 4.74e-2 & 1.26e-2 & 2.10e-1 & 19.10 \\
& Self-attention
& 2.20e-2 & 1.38e-2 & 6.59e-3 & 4.18e-2 & 1.70e-2 & 3.07e-2 & 3.61 \\
& Cross-attention
& \textbf{1.89e-2} & \textbf{1.36e-2} & \textbf{4.39e-3} & 4.69e-2 & 1.54e-2 & 3.41e-2 & 4.17 \\
& Hybrid-attention
& 3.63e-2 & 1.85e-2 & 5.89e-3 & \textbf{3.30e-2} & \textbf{1.10e-2} & \textbf{2.64e-2} & 3.89 \\
\midrule
\multirow{7}{*}{5}
& DeepONet
& 1.29e-2 & 1.64e-2 & 7.63e-2 & 1.56e0 & 6.23e-1 & 6.95e-1 & \textbf{0.83} \\
& Point-DeepONet
& 9.49e-3 & 2.16e-1 & 3.07e-1 & 9.38e-1 & 2.16e-1 & 4.23e-1 & {1.11} \\
& Transolver
& 8.12e-3 & \textbf{5.86e-3} & 6.25e-3 & 5.33e-2 & 2.20e-2 & 4.77e-2 & 18.18 \\
& Transolver+
& \textbf{6.39e-3} & 6.39e-3 & 6.50e-3 & 7.56e-2 & 2.51e-2 & 4.55e-2 & 19.47 \\
& Self-attention
& 6.69e-3 & 7.54e-3 & 7.28e-3 & 6.74e-2 & 2.91e-2 & 5.63e-2 & 3.89 \\
& Cross-attention
& 7.83e-3 & 7.13e-3 & \textbf{4.73e-3} & 8.11e-2 & 2.90e-2 & 5.83e-2 & 4.44 \\
& Hybrid-attention
& 7.37e-3 & 6.14e-3 & 6.23e-3 & \textbf{5.02e-2} & \textbf{2.06e-2} & \textbf{4.00e-2} & 4.17 \\
\bottomrule
\end{tabular}
\renewcommand{\arraystretch}{1}
\label{tabl_Rods_self_validation_errors}
\end{table}
\end{landscape}

\begin{table}[h!]
\centering
\caption{Redox flow battery problem: extrapolation relative $L_2$ errors for the streamwise velocity $v$ across different $nRods$ using the cross-attention ArGEnT model. Best (lowest) values in each row are in bold.}
\begin{tabular}{l|cccc}
\hline
Train / Test 
& nRods = 1 
& nRods = 3 
& nRods = 5 
& \makecell{nRods = 3 \\ (Curved Bound)} \\
\hline
Train: nRods = 1 
& \textbf{3.05e{-3}} & 3.58e{-1} & 4.59e{-1} & 3.71e{-1} \\

Train: nRods = 3 
& 3.83e{-2} & \textbf{1.54e{-2}} & 9.29e{-2} & 9.22e{-2} \\

Train: nRods = 5 
& 9.94e{-2} & 6.51e{-2} & \textbf{2.90e{-2}} & 1.44e{-1} \\
\hline
\end{tabular}
\label{tabl_Rods_cross_validation_errors}
\end{table}

\subsection{Jet Engine Bracket}\label{sec_bracket}
In the previous tests, all geometries were two-dimensional. To further evaluate the performance of the ArGEnT models on three-dimensional problems, we consider a jet engine bracket problem using the dataset released in \cite{park2026point}, which is constructed based on the DeepJEB geometries introduced in \cite{hong2025deepjeb}. The DeepJEB framework provides a family of brackets with non-parametric three-dimensional geometries that are difficult to represent using low-dimensional parameterizations. Several representative bracket geometries are shown in Fig.~\ref{Fig_jet_bracket_profiles} to provide an intuitive illustration.

\begin{figure}[htbp]
    \centering
    \includegraphics[width=15cm, trim=0cm 23.5cm 0cm 0cm, clip=true]{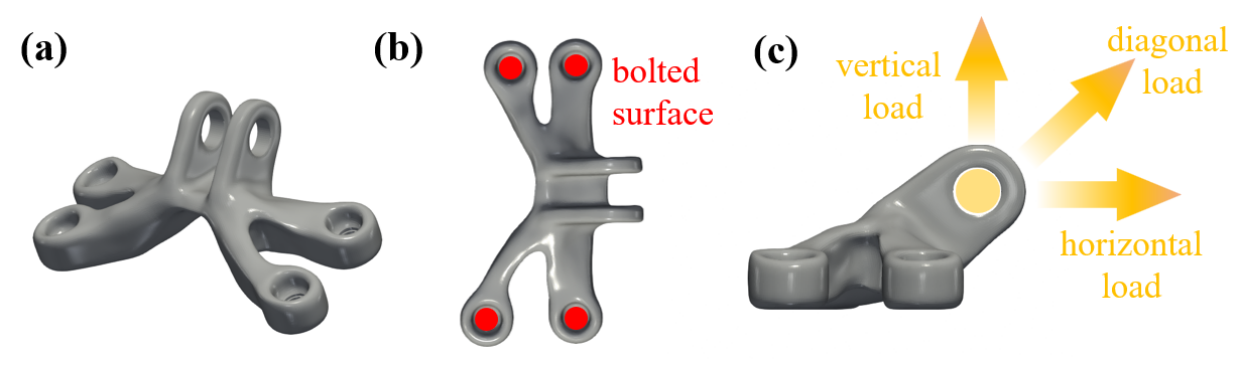}
    \caption{Jet engine bracket: (a) Isometric, (b) top, and (c) side views of the bracket geometry. Bolt locations are marked in red in the top view (b). Applied vertical, horizontal, and diagonal loading directions are illustrated by yellow arrows in (c).}
    \label{Fig_Bracket_geometry}
    
\end{figure}

The released dataset \cite{park2026point} includes displacement components $(u_x, u_y, u_z)$ and the von Mises stress field $\sigma_{vm}$ under three loading conditions: vertical, horizontal, and diagonal loads. These load cases are illustrated in Fig. \ref{Fig_Bracket_geometry}. The corresponding structural responses are obtained via nonlinear static finite element analysis. Details of the simulation setup and dataset generation procedure can be found in \cite{park2026point}. 

The full dataset contains 6,315 cases, with the number of mesh nodes ranging from approximately 120,000 to 380,000 across geometries. Following \cite{park2026point}, we use only 3,000 randomly selected cases for training and testing. To ensure a fair comparison between the ArGEnT models and their Point-DeepONet counterparts, we also sample 3,000 cases from the full dataset for model training and evaluation. Because of the random sampling procedure, the specific cases used here may differ from those used in \cite{park2026point}. In our sampled subset, there are 1,782 unique geometries, comprising 1,020 vertical, 985 horizontal, and 995 diagonal load cases. For comparison, \cite{park2026point} reports 1,785 unique geometries, with 969 vertical, 1,009 horizontal, and 1,022 diagonal load cases. We use 80\% of the samples for training and the remaining 20\% for testing. {For this problem, we consider only the cross-attention model because all three ArGEnT variants have already been compared in the preceding benchmark studies, whereas the present 3D case focuses on the scalability of cross-attention for large query sets.} To make the number of trainable parameters comparable between the cross-attention ArGEnT DeepONet and the Point-DeepONet\cite{park2026point}, we use narrower MLP widths (72 per layer) and a smaller attention embedding dimension (72) for the cross-attention model. As a result, the cross-attention ArGEnT DeepONet contains approximately 0.28 million parameters, comparable to the 0.25 million parameters of Point-DeepONet. For the cross-attention ArGEnT model, 5,000 points are sampled uniformly within a cuboid region enclosing all possible geometries. The signed distance function (SDF) values at these points are computed for each geometry and used as geometry inputs. To reduce the computational cost associated with large-mesh learning, 30,000 query points are sampled from the simulation mesh nodes for each geometry, and 5,000 points are randomly subsampled at each training step. The corresponding displacement components $(u_x, u_y, u_z)$ and von Mises stress ($\sigma_{vm}$) values at these points are then used as target outputs for training and evaluation of cross-attention ArGEnT.

The learned operator for the following jet engine bracket problem, taking the cross-attention ArGEnT as an example,  is therefore defined as
\begin{align}
\textbf{Cross-attention} \quad \mathcal{G} : (\tilde{\mathbf{x}}, \Omega, \pmb{\mu}) 
\longmapsto (u_x,\, u_y,\, u_z,\, \sigma_{vm} ) 
\end{align}
where $\pmb{\mu} =(mass, F_x, F_y, F_z)$ includes the bracket mass and the applied loads at $x, y, z$ directions. The definition of $\tilde{\mathbf{x}}$ and $\Omega$ follows Eqs. \eqref{eq_ArGEnT_trunk} and \eqref{eq_ArGEnT}. Since the solution variables exhibit different magnitudes, $z$-score normalization is applied to each variable, as defined in \ref{sec_norm}.

The prediction errors on the test set are summarized in Table~\ref{tab_jet_bracket}, which includes results from the standard DeepONet, PointNet, Point-DeepONet~\cite{park2026point}, and Transolver\cite{wu2024transolver}/Transolver+\cite{luo2025transolver++}. All models except Transolver and Transolver+ are configured to have comparable parameter counts (about 0.3 million) and are trained for 100,000 steps (approximately 1,333 epochs) with a batch size of 32. Transolver and Transolver+ follow their original configurations, with 0.9 million and 0.58 million parameters, respectively, and are trained with a batch size of 1 for 960,000 steps, corresponding to 400 epochs.

Compared with DeepONet, PointNet, and Point-DeepONet~\cite{park2026point}, cross-attention ArGEnT achieves consistently higher accuracy across all load conditions when trained with comparable numbers of trainable parameters and the same dataset size, albeit at increased computational cost. Compared with Transolver and Transolver+, cross-attention ArGEnT yields lower errors for the displacement fields $u_x$, $u_y$, and $u_z$ across all load conditions, while producing comparable errors for the von Mises stress field $\sigma_{vm}$. This behavior may be attributed to the much stronger local gradients in the von Mises stress field relative to the displacement fields, which pose a greater challenge for machine-learning-based surrogate models and may require a larger network to resolve accurately. We also investigate the scaling behavior of cross-attention ArGEnT by training it on the full dataset (6315 cases) with approximately three times more trainable parameters (0.87 million). This scaling leads to a substantial reduction in prediction errors, demonstrating strong scalability and highlighting the potential of cross-attention ArGEnT for realistic engineering applications.

\begin{remark}
\textit{In our experiments in Sections \ref{sec_lid}, \ref{sec_RFB}, and \ref{sec_bracket}, Point-DeepONet does not outperform DeepONet on the lid-driven cavity flow and redox flow battery problems, although it is reported to surpass DeepONet on the 3D jet engine bracket problem in the original study \cite{park2026point}. A key distinction is that the bracket setting does not admit an explicit low-dimensional geometric parametrization, whereas the cavity and battery problems considered here provide structured geometric descriptors that can be directly incorporated into the DeepONet branch network. When such geometric parameters are available, explicitly leveraging them may provide DeepONet with a representational advantage over a Point-DeepONet that relies solely on point-cloud representations. Nevertheless, Point-DeepONet retains a clear advantage in handling arbitrary geometries without requiring explicit parametrization.}
\end{remark}

\begin{figure}[htbp]
    \centering
    \includegraphics[width=15cm, trim=5cm 4.5cm 4cm 3cm, clip=true]{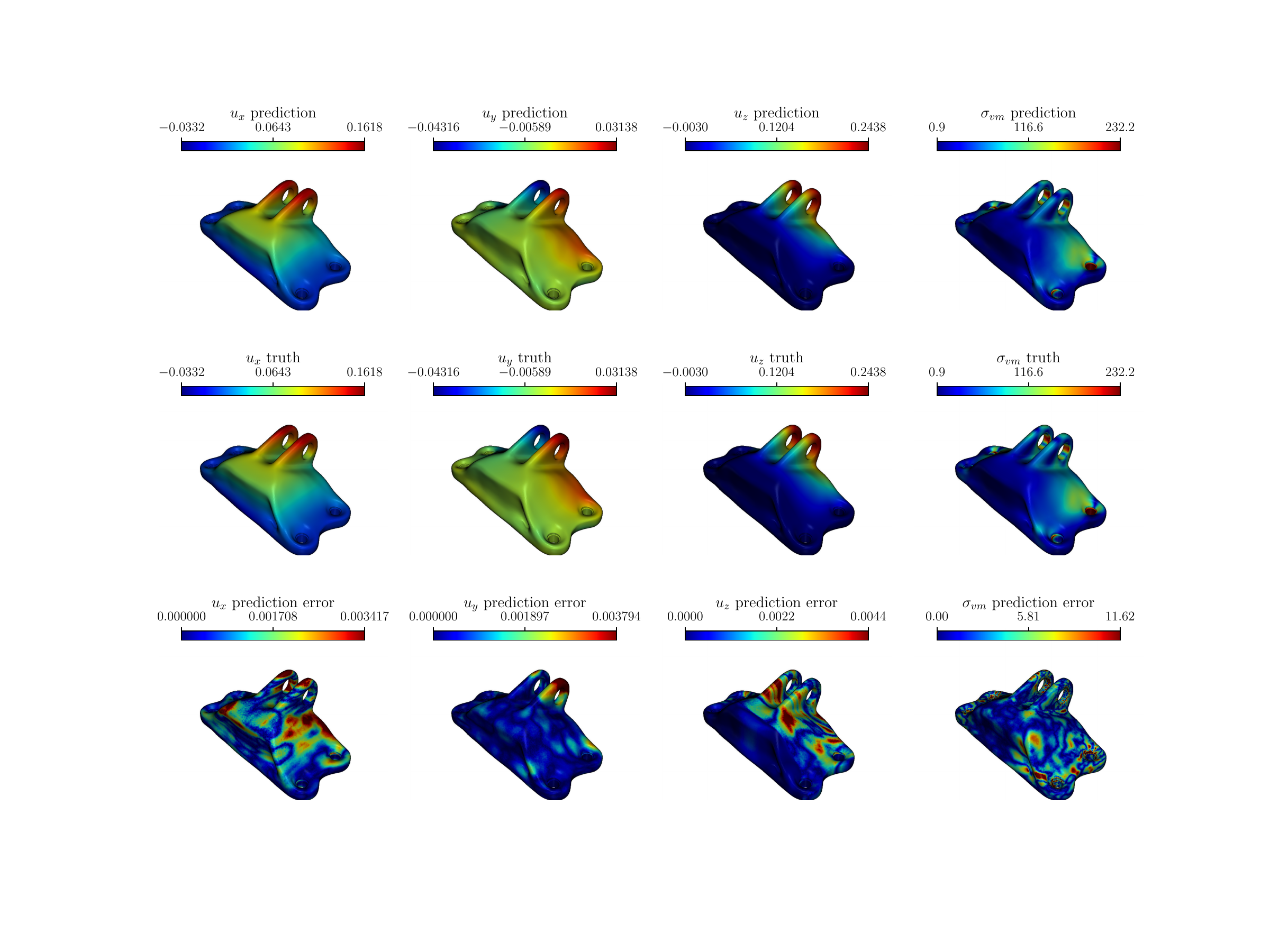}
    \caption{Jet engine bracket: contour plots of predicted structural response fields, ground truth and predicted absolute errors for a test case using the cross-attention transformer model. The error fields contain a small number of extreme outliers, especially for the von Mises stress field. For visualization purposes, values above the 99th percentile and below the 1st percentile are trimmed. Besides, the displacement components $(u_x,\, u_y,\, u_z)$ are shown in mm and the von Mises stress $\sigma_{vm}$ in MPa.}
    \label{Fig_bracket_casePrediction}    
\end{figure}

Figure~\ref{Fig_bracket_casePrediction} shows the predicted response fields and the corresponding absolute error distributions for a representative test case obtained using the cross-attention ArGEnT model (scaling-up setting). The model accurately captures the spatial distributions of both the displacement fields and the von Mises stress field. Compared with the displacement fields, the von Mises stress field exhibits much larger local gradients, which poses a greater challenge for machine-learning-based surrogate models. As shown in the figure, the prediction errors for the von Mises stress are noticeably larger in regions with strong local gradients. This behavior may be attributed to limitations in network expressivity as well as increased uncertainty originating from the underlying numerical simulation in high-gradient regions.

\begin{table}[htbp]
\centering
\caption{Jet engine bracket problem: Mean absolute errors of original field variables under different load directions. Best (lowest) values in each column are highlighted in bold. ``Cross-attention”, namely the cross-attention ArGEnT, uses parameter counts and dataset size comparable to Point-DeepONet; ``Cross-attention (scaling up)'' uses approximately 2.5× more parameters and a 2× larger dataset (the full dataset).}
\label{tab_jet_bracket}
\begin{tabular}{llcccc}
\toprule
Model & Load & $u_x$(mm) & $u_y$(mm)  & $u_z$(mm)  & $\sigma_{vm}$(MPa) \\
\midrule
\multirow{3}{*}{PointNet\cite{park2026point} }
& Vertical   & 0.008 & 0.003 & 0.013 & 11.666 \\
& Horizontal & 0.006 & 0.002 & 0.008 & 9.073  \\
& Diagonal   & 0.004 & 0.002 & 0.006 & 7.639  \\
\midrule
\multirow{3}{*}{DeepONet\cite{park2026point} }
& Vertical   & 0.026 & 0.007 & 0.048 & 21.842 \\
& Horizontal & 0.024 & 0.004 & 0.026 & 20.091 \\
& Diagonal   & 0.006 & 0.005 & 0.016 & 12.555 \\
\midrule
\multirow{3}{*}{Point-DeepONet\cite{park2026point} }
& Vertical   & 0.007 & 0.003 & 0.012 & 10.541 \\
& Horizontal & 0.005 & 0.002 & 0.006 & 7.935  \\
& Diagonal   & 0.003 & 0.002 & 0.005 & 7.090  \\
\midrule
\multirow{3}{*}{Transolver}
& Vertical   & 0.0085 & 0.0022 & 0.0154 & 7.4013 \\
& Horizontal & 0.0073 & 0.0015 & 0.0077 & 5.9132 \\
& Diagonal   & 0.0040 & 0.0013 & 0.0065 & 5.2288 \\
\midrule
\multirow{3}{*}{Transolver+}
& Vertical   & 0.0084 & 0.0024 & 0.0118 & 7.3257 \\
& Horizontal & 0.0099 & 0.0016 & 0.0071 & 5.9327 \\
& Diagonal   & 0.0047 & 0.0014 & 0.0055 & 5.1690 \\
\midrule
\multirow{3}{*}{\makecell{Cross-attention}}
& Vertical   & {0.0039} & {0.0018} & {0.0076} & {7.9454} \\
& Horizontal & {0.0033} & {0.0012} & {0.0047} & {6.5486} \\
& Diagonal   & {0.0017} & {0.0011} & {0.0031} &
{5.6045} \\
\midrule
\multirow{3}{*}{\makecell{Cross-attention\\(Scaling up)}}
& Vertical   & \textbf{0.0023} & \textbf{0.0013} & \textbf{0.0042} & \textbf{6.1865} \\
& Horizontal & \textbf{0.0020} & \textbf{0.0009} & \textbf{0.0026} & \textbf{4.9135} \\
& Diagonal   & \textbf{0.0012} & \textbf{0.0008} & \textbf{0.0021} & \textbf{4.5795} \\
\bottomrule
\end{tabular}
\end{table}

\section{Conclusion and Future Work}
In this work, we developed ArGEnT, a geometry-encoded transformer architecture designed to represent complex and irregular geometries through attention mechanisms. We introduced three ArGEnT variants—self-attention, cross-attention, and hybrid-attention—each providing a flexible and expressive way to encode geometric information. Although ArGEnT can be used as a standalone architecture, as demonstrated in the laminar airfoil case, we coupled it with DeepONet in the remaining examples to provide a representative testbed for systematically evaluating its effectiveness in geometry-aware surrogate modeling.

The resulting ArGEnT-based operator models were evaluated on a diverse set of problems, including laminar and turbulent airfoil flows, lid-driven cavity flow, a pore-scale redox flow battery system, and a 3D jet engine bracket problem. Across all test cases, models equipped with ArGEnT geometry encoders achieve significantly lower prediction errors than the standard DeepONet and other models reported in the literature. Moreover, ArGEnT exhibits strong generalization to unseen geometries that lie outside the expressive capacity of the geometry variations used during training, a regime in which the standard DeepONet fails. {Among the proposed variants, the cross-attention and hybrid-attention architectures consistently deliver the highest accuracy, while their query inputs do not necessarily require SDF values because geometry is encoded separately through the key-value inputs.} The self-attention model achieves comparable performance; however, only the cross-attention model shows robust behavior with respect to query-point sampling during inference.

{Despite these advantages, ArGEnT has several limitations in its current form. Although the linear-attention formulation improves scalability, the computational cost still increases with the numbers of geometry points, query points, and embedding dimensions, and geometries containing fine-scale features may require denser point-cloud representations. In addition, self-attention and hybrid-attention depend on the joint query-point set and are therefore most appropriate when the training and inference discretizations are fixed or similarly distributed. Cross-attention instead evaluates query points independently using a separate geometry representation, making it the preferred variant for arbitrary point evaluation. The present study is also limited to static geometries and does not explicitly address moving boundaries, nonstationary domains, or time-dependent PDEs.

{In future work, we plan to extend ArGEnT to moving and evolving geometries, time-dependent PDEs, and more complex multiphysics systems. We will also explore transfer learning and continual learning strategies to improve transferability and generalization across different problem settings. Finally, we will investigate the incorporation of physics-informed constraints into the ArGEnT DeepONet framework to further improve predictive accuracy and physical consistency. Because once the geometry representation is fixed, arbitrary query locations can be evaluated independently, this pointwise-independent query propagation may be particularly useful for dynamically sampled PDE residual points and boundary-condition points during physics-informed training. We view this property as a natural foundation for future physics-informed operator-learning work; physics-informed ArGEnT has not yet been demonstrated in the present manuscript.}}

\section*{Acknowledgments}
The research described in this paper was conducted as part of an interlaboratory collaboration involving Pacific Northwest National Laboratory (PNNL, Contract No. DE-AC05–76RL01830), Lawrence Livermore National Laboratory (LLNL, Contract No. DE-AC52-07NA27344), and Sandia National Laboratories (SNL, Contract No. DE-NA0003525) under the DOE Interlaboratory (IL) Laboratory Directed Research and Development (c) Pilot Program. Computational resources were provided by PNNL Research Computing. PNNL release number is PNNL-SA-219947. LLNL release number is LLNL-JRNL-2015467. 
Any subjective views or opinions that might be expressed in the paper do not necessarily represent the views of the U.S. Department of Energy or the United States Government.

\appendix
\section{Self-attention block}\label{appendix_self_attention}
The self-attention block consists of three main steps: query, key, and value projection. First, the input features are linearly transformed into three separate representations: queries, keys, and values. Next, the attention scores are computed by taking the dot product of the queries and keys, followed by a softmax operation to obtain the attention weights. Finally, the output is generated by taking a weighted sum of the values, where the weights are determined by the attention scores. The mathematical formulation of the self-attention mechanism is given by:
\begin{equation}
\text{Attention}(Q, K, V) = \text{softmax}\left(\frac{QK^T}{\sqrt{d_k}}\right)V
\end{equation}
where $Q \in R^{n \times d_q}$, $K \in R^{n \times d_k}$, and $V \in R^{n \times d_v}$ represent the query, key, and value matrices, respectively. The dimensions $d_q$, $d_k$, and $d_v$ denote the feature dimensions of the queries, keys, and values, respectively, while $n$ represents the number of input elements. $d_q$ should be equal to $d_k$ for the dot product operation to be valid, while $d_v$ can be different. For simplicity of implementation, we set $d=d_q = d_k = d_v$ throughout our models. The scaling factor $\sqrt{d_k}$ is used to prevent the dot products from becoming too large, which can lead to numerical instability during the softmax operation.

In practice, we can use linear attention mechanisms \cite{cao2021choose}, including the Fourier-type attention and Galerkin-type attention, to achieve higher accuracy and efficiency. The Fourier-type attention uses the Fourier-type integral transform to approximate the attention operation, while the Galerkin-type attention employs 
the Petrov–Galerkin-type projection to approximate the attention operation. The formulations of these two linear attention mechanisms are given by:
\begin{align}
\textbf{Fourier-type}&: \text{Attention}(Q, K, V) = (\widetilde{Q}\widetilde{K}^T){V}/n \\
\textbf{Galerkin-type}&: \text{Attention}(Q, K, V) = Q(\widetilde{K}^T\widetilde{V})/n,
\end{align}
where $\widetilde{\circ}$ denotes a trainable non-batch-based normalization operation, which can be implemented using the layer normalization. We note that the complexity of the standard, Fourier-type and Galerkin-type attentions have a complexity of $O(n^2d)$, $O(n^2d)$ and $O(nd^2)$, respectively. Since in our applications $d \ll n$, the Galerkin-type attention is more efficient than the other two attention mechanisms, and thus we use it in all our experiments unless otherwise specified.
 {Galerkin-type attention improves computational efficiency but is not equivalent to exact softmax attention and may be less flexible for highly localized interactions. The representation of sharp geometric features also depends on point-cloud resolution and embedding capacity.}
Note that the scaling factor $n$ in the Fourier-type and Galerkin-type attentions is the number of input elements for keys/values to maintain numerical stability.

\section{Cross-attention block}\label{appendix_cross_attention}
The cross-attention block is similar to the self-attention block, but it operates on two different input sequences: the query sequence and the key-value sequence. This allows the model to focus on relevant parts of the key-value sequence while processing the query sequence. The mathematical formulation of the cross-attention mechanism is given by:
\begin{align}
\textbf{Standard}&:\text{CrossAttention}(Q, K, V) = \text{softmax}\left(\frac{QK^T}{\sqrt{d_k}}\right)V \\
\textbf{Fourier-type}&: \text{CrossAttention}(Q, K, V) = (\widetilde{Q}\widetilde{K}^T){V}/n \\
\textbf{Galerkin-type}&: \text{CrossAttention}(Q, K, V) = Q(\widetilde{K}^T\widetilde{V})/n,
\end{align}
where $Q \in R^{m \times d_q}$, $K \in R^{n \times d_k}$, and $V \in R^{n \times d_v}$ represent the query, key, and value matrices, respectively. Similar to the self-attention block, we set $d=d_q = d_k = d_v$ for simplicity, and also the scaling factor $\sqrt{d_k}$ in the standard attention is used to maintain numerical stability.

\section{Rotary Position Embeddings (RoPE)}\label{appendix_RoPE}
The attention mechanism in Transformer architectures is inherently position-agnostic, since the dot-product operation between query and key vectors depends only on their feature representations. To incorporate positional information, different strategies have been proposed, such as absolute sinusoidal embeddings~\cite{vaswani2017attention} and learnable position vectors. However, these methods encode absolute positions rather than relative ones, which may limit generalization to unseen sequence lengths or shifted spatial coordinates.  

Rotary Position Embeddings (RoPE) \cite{su2024roformer} address this limitation by encoding positional information through rotations applied directly to the query and key embeddings. Instead of adding sinusoidal terms, RoPE defines a mapping
\begin{equation}
\psi(\mathbf{q}_i, x_i) = \Theta(x_i)\mathbf{q}_i ,
\end{equation}
where $x_i$ is the coordinate of the $i$-th token (or spatial point), $\mathbf{q}_i \in \mathbb{R}^d$ is its embedding, and $\Theta(x_i)$ is a block-diagonal rotation matrix:
\begin{equation}
\Theta(x_i) = \text{Diag}(\mathbf{R}_1, \mathbf{R}_2, \ldots, \mathbf{R}_{d/2}), 
\quad
\mathbf{R}_l = 
\begin{bmatrix}
\cos(\lambda x_i \theta_l) & -\sin(\lambda x_i \theta_l) \\
\sin(\lambda x_i \theta_l) & \cos(\lambda x_i \theta_l)
\end{bmatrix}.
\end{equation}

Here:  
\begin{itemize}
    \item $\lambda$ is the wavelength of the spatial or temporal domain (e.g., $\lambda = 2048$ for a discretized domain with 2048 grid points).  
    \item $\theta_l$ is the frequency assigned to the $l$-th embedding pair, commonly defined as
    \begin{equation}
    \theta_l = 10000^{-\frac{2(l-1)}{d}}, 
    \quad l \in \{1, 2, \dots, d/2\},
    \end{equation}
    following the frequency scaling in the original sinusoidal positional encoding~\cite{vaswani2017attention}.
\end{itemize}

A key property of RoPE is that the attention score depends only on the \emph{relative displacement} between positions:
\begin{equation}
\psi(\mathbf{q}_i, x_i)^\top \psi(\mathbf{k}_j, x_j) 
= \mathbf{q}_i^\top \Theta(x_i - x_j)\mathbf{k}_j,
\end{equation}
where $\mathbf{k}_j$ is the key embedding at coordinate $x_j$. This ensures that the attention mechanism is translation-invariant, making it naturally suited for tasks where relative, rather than absolute, positions are most informative.  

{RoPE can be extended to higher-dimensional domains by partitioning the embedding dimensions into coordinate-specific groups and applying an independent rotary transformation to each spatial axis. For a D-dimensional coordinate, the embedding is divided as evenly as possible among the D axes. Each group is rotated using the corresponding coordinate, and the rotated groups are then concatenated. When the embedding dimension is not exactly divisible by D, only the largest evenly divisible subset is used for the rotary transformations, while the remaining dimensions are left unchanged.}

\section{Geometry distributions} \label{sec_geometry_distributions}
\begin{figure}[htbp]
    \centering
    \subfigure[All airfoil profiles used in the laminar airfoil flow problem]{
    \includegraphics[width=15cm, trim=4.5cm 14cm 3.5cm 2cm, clip=true]{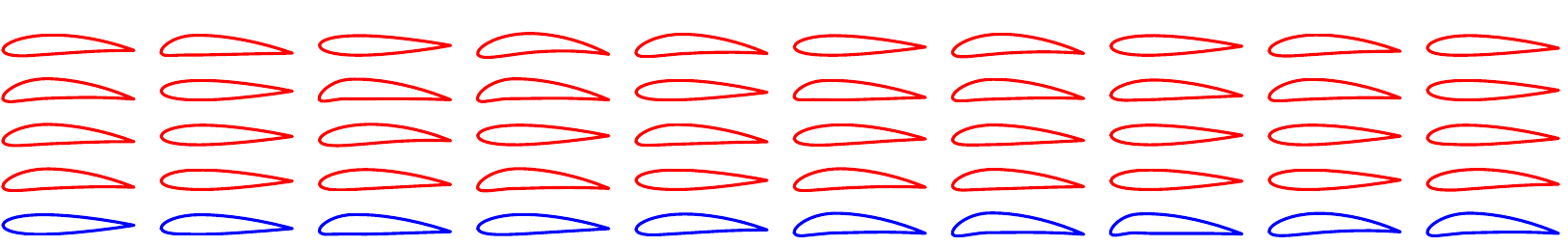}
    }

    \subfigure[Partial airfoil profiles used in the turbulent airfoil flow problem]{    
    \includegraphics[width=15cm, trim=4.5cm 2cm 3.5cm 2cm, clip=true]{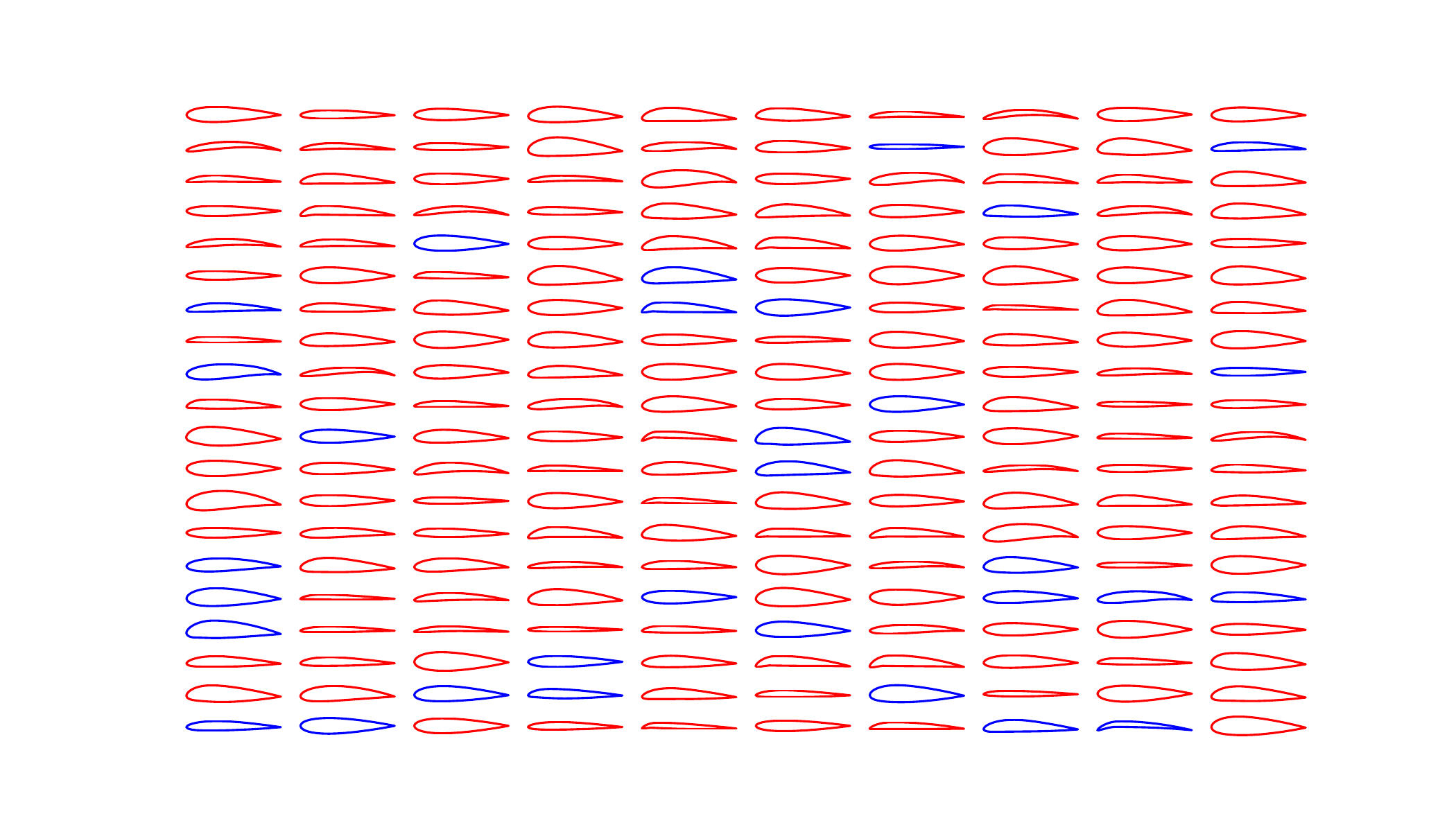}
    }
    \caption{Airfoil profiles used in the laminar and turbulent airfoil flow problems. The red airfoils are in the training set, while the blue airfoils are in the test set.}
    \label{Fig_airfoil_profiles}    
\end{figure}

Figure \ref{Fig_airfoil_profiles} visualizes the distribution of airfoil geometries used in the laminar and turbulent airfoil flow problems, illustrating the geometric variability covered by the training and test sets.

\begin{figure}[htbp]
    \centering
    \subfigure[Partial geometry profiles used in the lid-driven cavity flow problem]{
    \includegraphics[width=15cm, trim=4.5cm 8cm 3.5cm 2cm, clip=true]{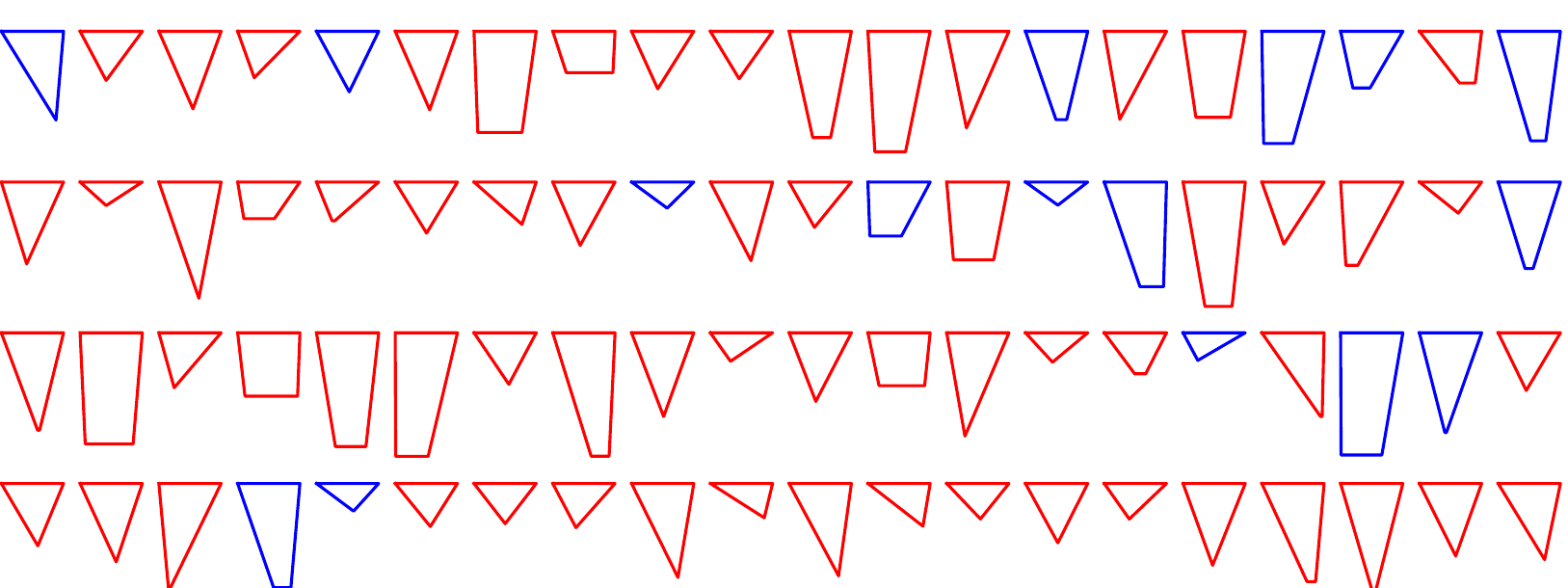}
    }

    \subfigure[Partial geometry profiles used in Redox flow battery problem]{    
    \includegraphics[width=15cm, trim=4.5cm 2cm 3.5cm 2cm, clip=true]{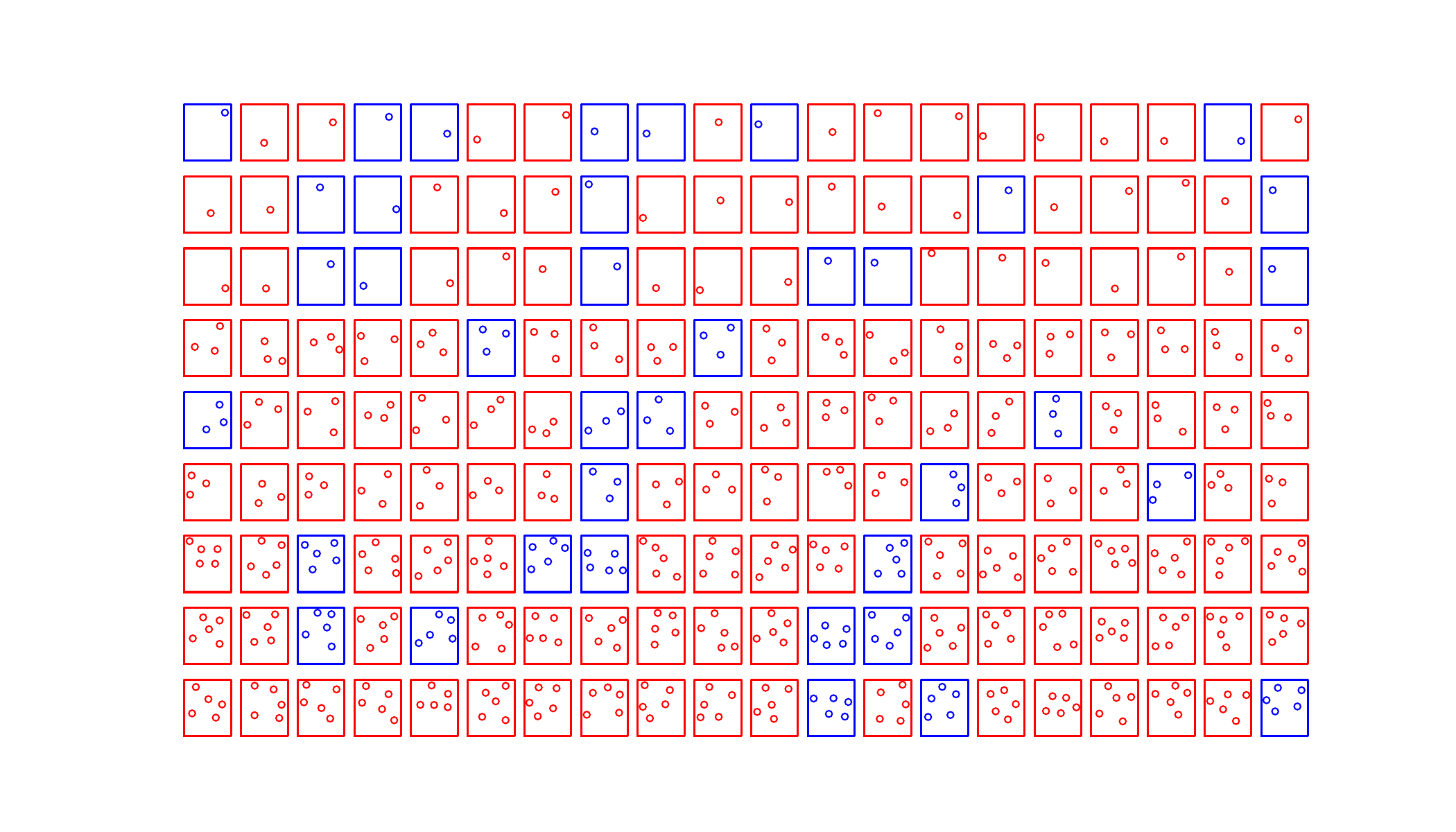}
    }
    \caption{Geometry profiles used in the lid-driven cavity flow and Redox flow battery problems. The red geometries are in the training set, while the blue geometries are in the test set.}
    \label{Fig_LidDriven_Rods_profiles}    
\end{figure}

\begin{figure}[htbp]
    \centering
    \includegraphics[width=15cm, trim=4.5cm 0cm 3.5cm 2cm, clip=true]{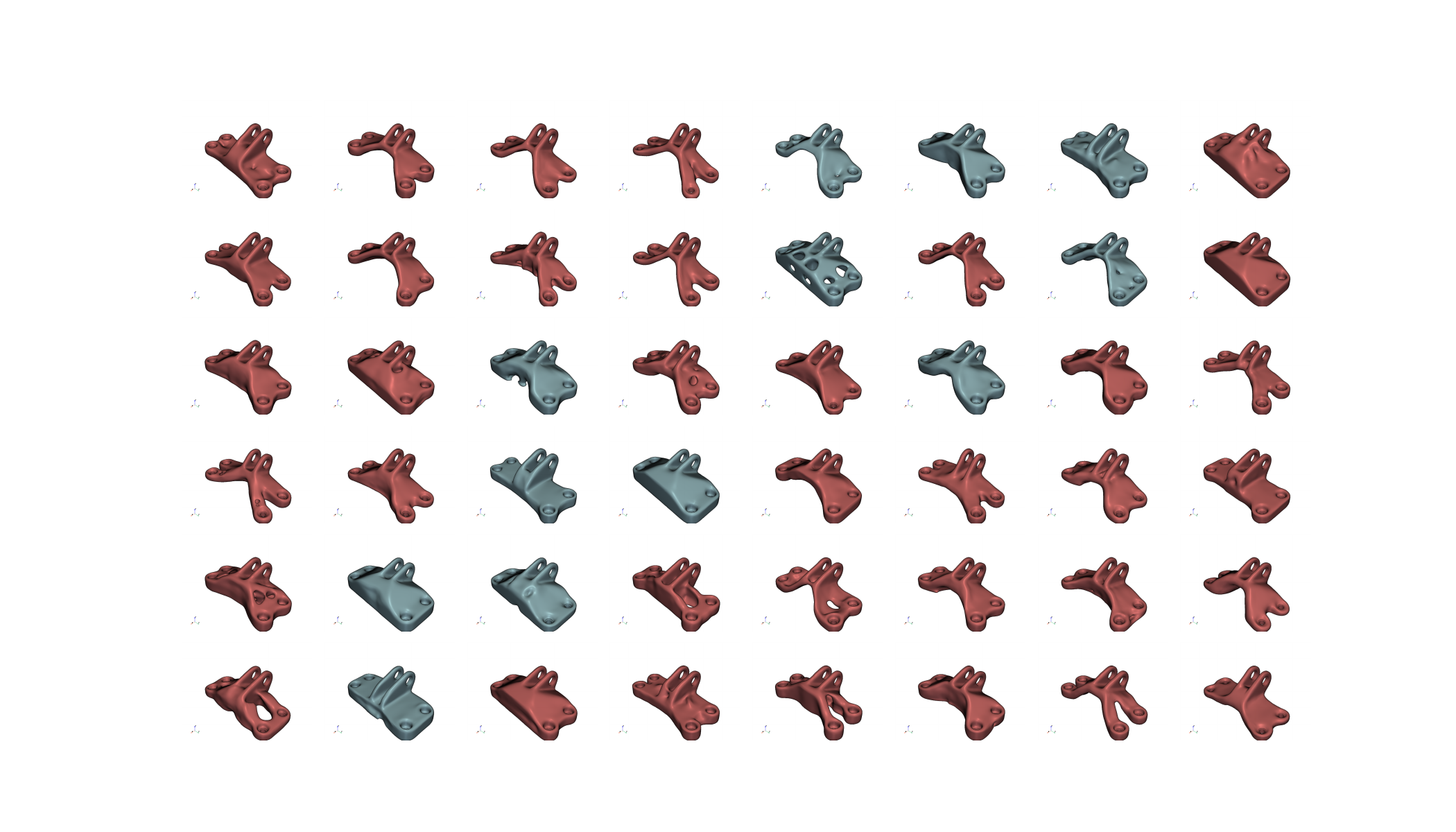}
    \caption{Partial geometries used in the jet engine bracket problem. The red geometries are in the training set, while the blue geometries are in the test set.}
    \label{Fig_jet_bracket_profiles}    
\end{figure}

Figures \ref{Fig_LidDriven_Rods_profiles} and \ref{Fig_jet_bracket_profiles} visualize the distribution of geometries used in the lid-driven cavity flow, Redox flow battery problems and jet engine bracket problem, illustrating the geometric variability covered by the training and test sets.

\section{Redox flow battery model} \label{sec_RFB_model}
The Redox flow battery is modeled as a coupled multiphysics system involving fluid flow, species transport, charge transport, and interfacial electrochemical reactions. For testing purpose, a simplified two-dimensional (2D) pore-scale redox flow battery model is adopted for data generation. The model is not intended to represent a full cell-scale device; instead, it serves as a controlled testbed for ArGEnT development and testing its capability for geometry-dependent generalization. The governing equations are summarized as follows.

\begin{itemize}
\item \textbf{Fluid transport.}  
The electrolyte flow is governed by the conservation of mass and momentum,
\begin{equation}
\nabla \cdot (\mathbf{u}) = 0,
\end{equation}
\begin{equation}
\nabla \cdot (\rho \mathbf{u}\mathbf{u})
= - \nabla p
+ \mu \nabla \cdot
\left( \nabla \mathbf{u} + (\nabla \mathbf{u})^{T} \right)
+ \rho \mathbf{g},
\end{equation}
where $\rho$ is the fluid density, $\mathbf{u}$ is the velocity vector, $p$ is the pressure, and $\mu$ denotes the electrolyte dynamic viscosity.

\item \textbf{Species transport.}  
The transport of ionic species $j$ follows an advection--diffusion--migration equation,
\begin{equation}
\mathbf{u} \cdot \nabla C_j
= - D_j \nabla^2 C_j
+ \nabla \cdot
\left(
\frac{z_j D_j C_j}{RT} \nabla \phi_e
\right),
\end{equation}
where $C_j$ is the concentration, $D_j$ is the diffusion coefficient, $z_j$ is the ionic charge number, $R$ is the universal gas constant, $T$ is the temperature, and $\phi_e$ is the electrolyte-phase electric potential.

\item \textbf{Charge transport.}  
Charge transport is modeled separately in the solid electrode and electrolyte phases. To simplify the model, the solid electrode potential is assumed to be spatially uniform and set to zero:
\begin{equation}
\phi_s=0
\end{equation}
Under this assumption, the solid-phase charge transport equation is not solved explicitly.

In the electrolyte phase, charge conservation is enforced as:
\begin{equation}
\nabla \cdot \left(
\kappa_{eff} \nabla \phi_e
+ F \sum_j z_j D_j \nabla C_j
\right) = 0,
\end{equation}
where $\kappa_s$ and $\kappa_{eff}$ denote the electrolyte conductivities, respectively, and $F$ is Faraday’s constant.

\item \textbf{Interfacial Electrochemistry.}  
Electrochemical reactions occur at the resolved electrode–electrolyte interface and are described by the Butler--Volmer equation in terms of local interfacial current density:
\begin{equation}
i
= F k_0
\left( \frac{C_{\mathrm{ox}}}{C_{\mathrm{ref}}} \right)^{\alpha_{\mathrm{an}}}
\left( \frac{C_{\mathrm{red}}}{C_{\mathrm{ref}}} \right)^{\alpha_{\mathrm{cat}}}
\left[
\exp\!\left( \frac{\alpha_{\mathrm{an}} F \eta}{RT} \right)
-
\exp\!\left( -\frac{\alpha_{\mathrm{cat}} F \eta}{RT} \right)
\right],
\end{equation}

where $i$ is the interfacial current density, $k_0$ is the reaction rate constant, $C_{ox}$ and $C_{red}$ are the oxidized and reduced species concentrations at the interface, $C_{ref}$ is a reference concentration, and $\alpha_{an}$ and $\alpha_{cat}$ are the anodic and cathodic transfer coefficients.

\item \textbf{Overpotential definition}  
With the solid-phase potential fixed at zero, the local overpotential is defined as
\begin{align}
\eta &= \phi_s - \phi_e - E_{\mathrm{eq}}, \\
     &= - \phi_e - E_{\mathrm{eq}},
\end{align}
where $\phi_e$ is the electrolyte-phase electric potential and $E_{\mathrm{eq}}$ is the local equilibrium (Nernst) potential.

\item \textbf{Equilibrium potential for the DHPS redox couple}
For the DHPS-based anolyte operating in alkaline electrolyte \cite{zeng2022characterization}, the two-electron redox reaction is written as
\begin{equation}
\mathrm{R} + 2\,\mathrm{H_2O} + 2e^- \;\rightleftharpoons\; \mathrm{R2H} + 2\,\mathrm{OH^-}.
\end{equation}
where $R$ and $R2H$ denote the oxidized and reduced forms of the DHPS redox couple, respectively.
The corresponding equilibrium potential is given by the Nernst equation
\begin{equation}
E_{\mathrm{eq}}
=
E_0
+
\frac{RT}{2F}
\ln\!\left(
\frac{C_{\mathrm{R}}\, c_{\mathrm{H_2O}}^{\,2}}
     {C_{\mathrm{R2H}}\, c_{\mathrm{OH^-}}^{\,2}}
\right),
\end{equation}
where $C_{\mathrm{R}}$ and $C_{\mathrm{R2H}}$ are the concentrations of the oxidized and reduced DHPS species, respectively, $c_{\mathrm{H_2O}}$ and $c_{\mathrm{OH^-}}$ are the local concentrations of water and hydroxide ions, and $E_0 = -1.06~\mathrm{V}$ is the formal redox potential of the DHPS redox couple (vs.\ Ag/AgCl).

\end{itemize}

\section{Influence of signed distance function (SDF) as an additional input feature}\label{sec_SDF}

To investigate the impact of incorporating the signed distance function (SDF) as an additional input feature, we conduct supplementary experiments on the lid-driven cavity flow and redox flow battery problems. For both the standard DeepONet and the attention-based ArGEnT DeepONet models, we compare the prediction accuracy obtained with and without including the SDF in the trunk network inputs.
For the without SDF setting in the cross-attention and hybrid-attention models, the SDF is removed only from the query inputs, while it is retained in the key–value inputs, since the geometry is implicitly encoded through the SDF distribution in the key–value representations. In contrast, for the self-attention model, the SDF is completely removed from all input features in the without SDF setting to ensure the validity and consistency of the self-attention mechanism.

\begin{figure}[htbp]
    \centering
    \includegraphics[width=14cm, trim=2cm 1.5cm 2cm 2cm, clip=true]{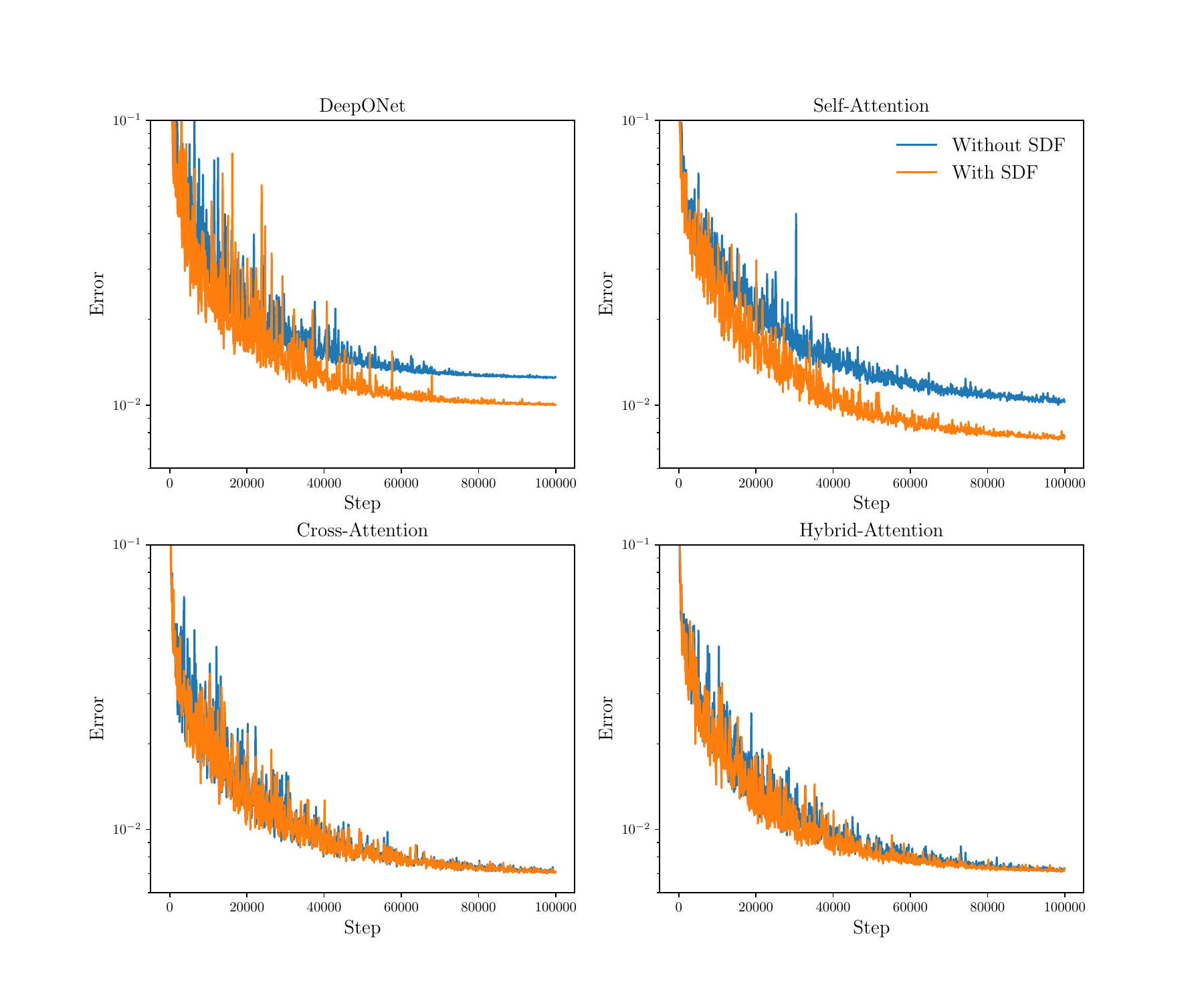}
    \caption{Lid-driven cavity flow: test errors versus training steps for different models. SDF denotes the use of signed distance function as an additional input feature.}
    \label{Fig_LidDriven_testErrors_step}
\end{figure}

\begin{figure}[htbp]
    \centering
    \includegraphics[width=14cm, trim=1.5cm 1.5cm 2cm 2cm, clip=true]{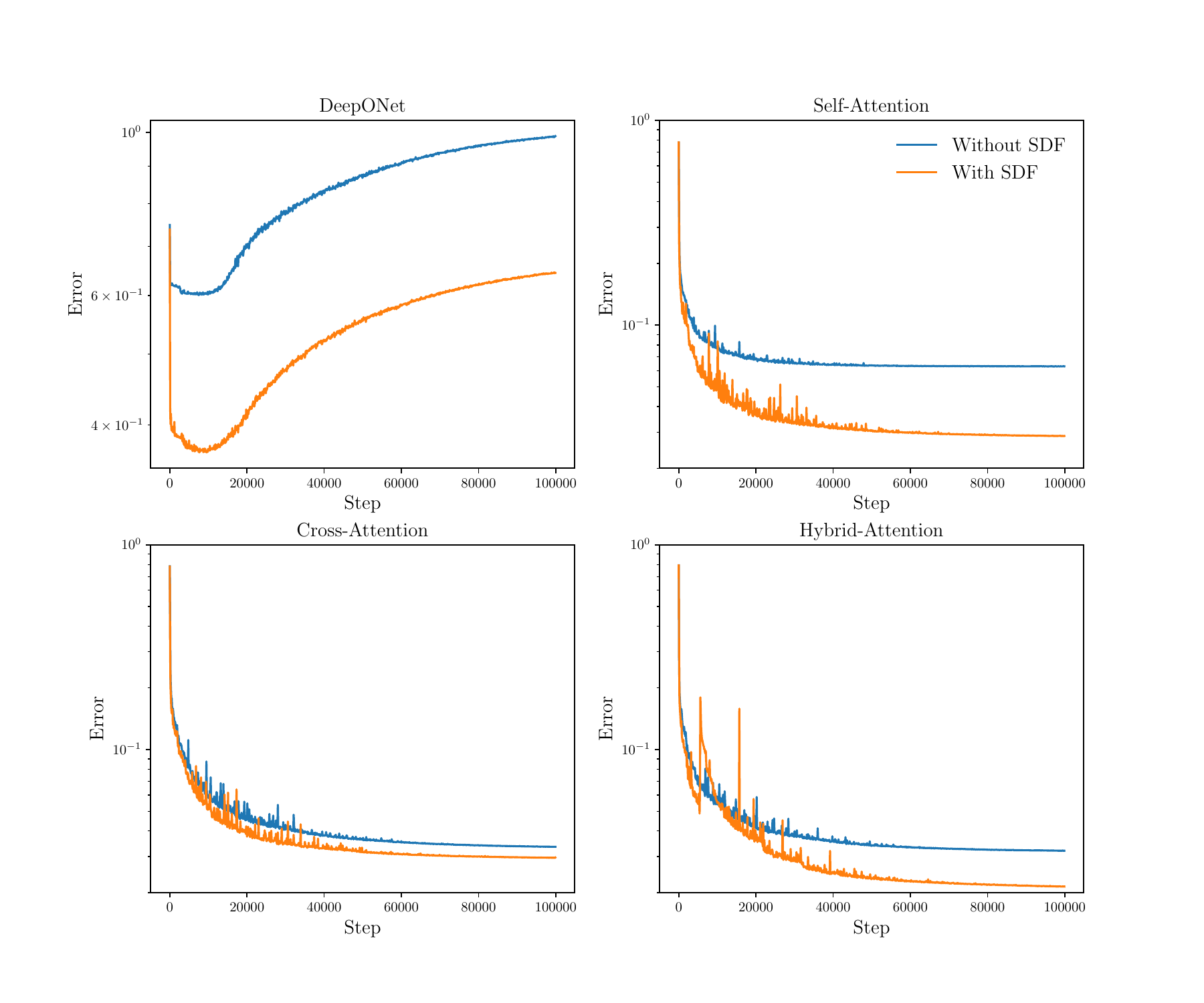}
    \caption{Redox flow battery with 5 rods: test errors versus training steps for different models. SDF denotes the use of signed distance function as an additional input feature.}
    \label{Fig_Rods_testErrors_step}
\end{figure}

The test errors as a function of training steps for the lid-driven cavity flow and redox flow battery problems are shown in Figures~\ref{Fig_LidDriven_testErrors_step} and \ref{Fig_Rods_testErrors_step}, respectively, and the final test errors are summarized in Table~\ref{tab_test_error_lid_rods}. The results indicate that incorporating the SDF improves prediction accuracy for the standard DeepONet and the self-attention ArGEnT DeepONet models. This improvement can be attributed to the explicit geometric information provided by the SDF, which facilitates more accurate modeling of the influence of complex geometries on the solution fields.
In contrast, for the cross-attention and hybrid-attention ArGEnT DeepONet models, the inclusion of the SDF does not lead to noticeable accuracy gains. This behavior arises because these architectures can effectively extract geometric information directly from the geometry representations in the key and value inputs, making the explicit SDF input less essential. This observation highlights the capability of cross-attention and hybrid-attention mechanisms to learn complex geometric relationships from data alone, thereby reducing reliance on auxiliary geometric features. Such a property is particularly advantageous in practical applications where the SDF or its associated quantities are difficult to obtain. For example, in physics-informed training settings that involve gradients of the solution fields with respect to spatial coordinates, incorporating SDF gradients may introduce additional computational overhead.

\begin{table}[htbp]
\centering
\caption{Test errors for the Lid-driven cavity flow and the Redox flow battery problem with and without the signed distance function (SDF) as an additional input feature.}
\label{tab_test_error_lid_rods}
\begin{tabular}{lcccc}
\toprule
Model 
& \multicolumn{2}{c}{Lid-driven cavity} 
& \multicolumn{2}{c}{Redox flow battery} \\
\cmidrule(lr){2-3} \cmidrule(lr){4-5}
& Without SDF & With SDF & Without SDF & With SDF \\
\midrule
DeepONet         
& 0.0125 & 0.0100 
& 0.9884 & 0.6442 \\
Self-Attention   
& 0.0103 & 0.0078 
& 0.0629 & 0.0287 \\
Cross-Attention  
& 0.0071 & 0.0071 
& 0.0335 & 0.0297 \\
Hybrid-Attention 
& 0.0073 & 0.0073 
& 0.0321 & 0.0214 \\
\bottomrule
\end{tabular}
\end{table}

\section{Normalization}\label{sec_norm}

For the turbulent airfoil flow problem in Section~\ref{sec_turbulent_airfoil}, the input variables $(x,y,d)$ are normalized with the chord length of the airfoil, which is set to 1. The freestream velocity $(U_{\inf}, V_{\inf})$ as well as the output variables are all normalized in the sense of $z$ score normalization, i.e., $\hat{z} = (z-\mu_z)/\sigma_z$, where $\mu_z$ and $\sigma_z$ denote the mean and standard deviation, which are  given in Table~\ref{tab_turbulent_airfoil_normalization}.

\begin{table}[htbp]
\centering
\caption{Normalization parameters ($z$ score) for the turbulent airfoil flow problem.}
\label{tab_turbulent_airfoil_normalization}
\begin{tabular}{lcccccc}
\toprule
Var  & $u$ & $v$ & $p$ & $\nu_t$ & $U_{\inf}$ & $V_{\inf}$\\
\midrule
Mean & 42.50 & 9.87 & $-456.50$ & $7.97\times10^{-4}$ & 61.70 & 4.93\\
Std  & 29.73 & 31.01 & $2921.73$ & $2.91\times10^{-3}$ & 17.72 & 6.33\\
\bottomrule
\end{tabular}
\end{table}

For Redox flow battery problem in \ref{sec_RFB_model}, the input variables $(x,y,d)$ are normalized with the channel width $100 \mu m$. The input velocity is normalized with the mean inlet velocity 5 mm/s.
The output variables are all normalized in the sense of $z$-score normalization, i.e., $\hat{z} = (z-\mu_z)/\sigma_z$, where $\mu_z$ and $\sigma_z$ denote the mean and standard deviation of the variable $z$, which are manually given in Table~\ref{tab_RFB_normalization}. 

\begin{table}[htbp]
\centering
\caption{Normalization parameters ($z$ score) for the redox flow battery problem.}
\label{tab_RFB_normalization}
\begin{tabular}{lcccccc}
\toprule
Var & $\phi^{-}_{e}$ & $\eta_{-}$ & $c_R$ & $u$ & $v$ & $p$ \\
\midrule
Mean & 0.8963 & 0.0458 & 825 & 0 & 0 & 1.54 \\
Std  & 0.0232 & 0.0231 & 33 & $v_{\mathrm{in}}$ & $v_{\mathrm{in}}$ & 1.80 \\
\bottomrule
\end{tabular}
\end{table}

For jet engine bracket problem in \ref{sec_RFB_model}, the input variables $(x,y,z, d)$, the non-parametric parameters $\mu=(mass, F_x, F_y, F_z)$, and the output variables $(u_x, u_y, u_z, \sigma_{vm})$ are all normalized in the sense of $z$ score normalization, i.e., $\hat{z} = (z-\mu_z)/\sigma_z$, where $\mu_z$ and $\sigma_z$ denote the mean and standard deviation of the variable $z$, which are manually given in the Table~\ref{tab_bracket_normalization}.

\begin{table}[htbp]
\centering
\caption{Normalization parameters ($z$ score) for the jet engine bracket problem.}
\label{tab_bracket_normalization}
\begin{tabular}{lcccccc}
\toprule
Var & $x$ & $y$ & $z$ & $d$ & $mass$ & $F_x$ \\
\midrule
Mean & 15.02 & -54.30 & 15.49 & -1.26 & 0.63 & -13.54 \\
Std  & 23.12 & 47.16 & 13.84 & 1.77 & 0.63 & 16.49  \\
\toprule
Var &  $F_y$ & $F_z$ & $u_x$ & $u_y$ & $u_z$ & $\sigma_{vm}$ \\
\midrule
Mean & 8.50 & 13.06 & -8.26e-3 & -5.81e-4 & 0.04 & 38.37 \\
Std  & 16.49 & 11.66 & 7.66e-2 & 1.53e-2 & 0.12 & 44.80 \\
\bottomrule
\end{tabular}
\end{table}

\bibliographystyle{elsarticle-num}
\bibliography{reference.bib}

\end{document}